%% file: main.tex
\documentclass{article}

\PassOptionsToPackage{numbers,sort&compress}{natbib}
\usepackage[main,final]{neurips_2026}

\usepackage{booktabs}
\usepackage{amsmath}
\usepackage{amssymb}
\usepackage{amsfonts}
\usepackage{array}
\usepackage{tabularx}
\usepackage[hidelinks,pdftitle={Routing Subspaces: Auditing Evaluation-to-Deployment Mismatch in Fine-Tuned Language Models},pdfauthor={Phongsakon Mark Konrad, Toygar Tanyel, Serkan Ayvaz}]{hyperref}
\usepackage{etoc}
\usepackage{fontawesome5}
\usepackage{url}
\usepackage{microtype}
\usepackage{xcolor}
\usepackage{graphicx}
\usepackage{subcaption}
\usepackage{float}
\usepackage{tikz}
\usetikzlibrary{arrows.meta, calc, decorations.pathmorphing, positioning, shapes.geometric, backgrounds, fit, shadows.blur}

\definecolor{tolblue}{HTML}{4477AA}
\definecolor{tolred}{HTML}{EE6677}
\definecolor{tolgreen}{HTML}{228833}
\definecolor{tolyellow}{HTML}{CCBB44}
\definecolor{tolgrey}{HTML}{888888}
\definecolor{tollight}{HTML}{D9D9D9}
\definecolor{tolink}{HTML}{222222}
\definecolor{tolaccent}{HTML}{AA3377}
\colorlet{evalcol}{tolblue}
\colorlet{depcol}{tolred}
\colorlet{dirCol}{tolink}
\colorlet{ghost}{tollight}
\colorlet{midcol}{tolgreen}

\newcommand{\routetablestyle}{\small\setlength{\tabcolsep}{4pt}\renewcommand{\arraystretch}{1.1}}
\newcommand{\hd}[2]{\shortstack{\textbf{#1}\\\textbf{#2}}}
\newcommand{\upbetter}{\ensuremath{\uparrow}}
\newcommand{\downbetter}{\ensuremath{\downarrow}}

\newcommand{\EvalFr}{\mathrm{E}}
\newcommand{\DeployFr}{\mathrm{D}}
\newcommand{\hvec}{\mathbf{h}}
\newcommand{\dvec}{\mathbf{d}}
\newcommand{\dhat}{\hat{\mathbf{d}}}
\newcommand{\muvec}{\boldsymbol{\mu}}
\newcommand{\Wmat}{\mathbf{W}}
\newcommand{\Imat}{\mathbf{I}}
\newcolumntype{Y}{>{\raggedright\arraybackslash}X}
\newcolumntype{R}[1]{>{\raggedright\arraybackslash}p{#1}}

\title{Routing Subspaces:\\
       Auditing Evaluation-to-Deployment Mismatch in Fine-Tuned Language Models}

\author{%
  Phongsakon Mark Konrad$^{1}$
  \And
  Toygar Tanyel$^{2}$
  \And
  Serkan Ayvaz$^{1}$
  \AND
  \textnormal{$^{1}$Centre for Industrial Software, University of Southern Denmark, Sønderborg, Denmark} \\
  $^{2}$Promake, Newark, DE, USA \\
  Correspondence to: Serkan Ayvaz \texttt{<seay@mmmi.sdu.dk>} \\[4pt]
  {\large\href{https://github.com/aiml-sdu/routing-subspaces}{\faGithub}\quad\href{https://aiml-sdu.github.io/routing-subspaces-page/}{\faGlobe}}
}

\begin{document}

\etocdepthtag.toc{main}

\maketitle

\begin{abstract}
Safety evaluations often assume that behavior observed during testing reflects behavior in ordinary use, but fine-tuning can break this assumption. A checkpoint can appear fixed under evaluation-style prompts while the same behavior persists under ordinary-use prompts. Output scores reveal this mismatch but do not locate it. We investigate whether the distinction is encoded in a stable internal site and introduce an audit that fits a paired activation contrast at a path-patching-informed mid-depth window, then modifies the resulting coordinate on held-out prompts. The intervention reduces the evaluation-to-deployment gap by more than $0.12$ in ten of twelve model--behavior settings and closes it (post-intervention gap below $0.06$) in four, across four full-matrix instruction-tuned model instances; among the eight settings with $n{\geq}120$ paired questions, six reduce and one closes. A fifth model supports localization and edit-provenance checks, and in the twelve settings deployment-framed rates change by at most $6.1$pp. On retrained checkpoints, random directions at the same dose do not reproduce the effect, and off-window controls do so only in part. The strongest literal evidence for an evaluation-versus-deployment distinction comes from sandbagging, whose prompt framing is explicitly evaluative; sycophancy and refusal test framing-contingent task distinctions. The two weak cells, both sycophancy, show where a single-coordinate audit at a fixed window is not sufficient. The audit is a non-adaptive, white-box diagnostic for controlled fine-tuned checkpoints, not a training-time defense, a guarantee of deployment safety, or evidence about naturally arising failures.
\end{abstract}

\section{Introduction}
\label{sec:intro}

Fine-tuned language models are routinely evaluated in settings that differ from those in which they will be deployed. A checkpoint can answer an evaluation-style prompt incorrectly while answering the same content correctly as a user query. It can withhold a correct answer when a request declares malicious intent while giving it when the same question arrives with a benign one. Related model-organism studies construct conditional behaviors such as backdoors, password-locked capabilities, and training/out-of-training alignment faking \citep{hubinger2024sleeper,greenblatt2024passwordlocked,greenblatt2024alignmentfaking}, but do not identify where the distinction is represented or how to probe it after training.

Output scores reveal the mismatch but do not localize it. Three questions remain open: whether the distinction is written into a stable internal site, whether it generalizes to held-out prompts, and whether a flat probe means the behavior is absent or the probe missed it. Internal representations can carry latent knowledge not settled by surface behavior \citep{burns2022latentknowledge}. We ask whether a fine-tuned checkpoint leaves a comparable internal trace of the framing distinction.

The audit target is narrower than a general steering vector and more operational than a descriptive linear separator. It is tied to the fine-tuned checkpoint and to paired evaluation-framed and deployment-framed prompts, and tested by an intervention on held-out examples rather than by linear separability alone. If the same intervention also moves random directions, wrong layers, or ordinary semantic contrasts, the audit has learned little about framing. If it closes the gap while those controls stay flat, the mismatch has a concrete internal target.

Prior work makes this question plausible but does not answer it. Activation-steering work shows that many behaviors can be moved by activation-space directions or low-dimensional inference-time interventions \citep{subramani2022steering,li2023iti,arditi2024refusal,rimsky2024caa,zou2023representation,turner2023activation}. Circuit-localization work shows how to trace behavior through internal components \citep{wang2023ioi,conmy2023automated,goldowskydill2023pathpatching}. Evaluation and situational awareness can be measured, probed, or steered in individual models \citep{berglund2023situational,laine2024sad,needham2025evaluated,nguyen2025evalawareness,hua2025evalawareness}. What remains unclear is whether fine-tuning leaves a comparable audit target: localizable at a path-patching-informed site, testable against matched controls, and typed when it fails.

Throughout the main intervention matrix, ``twelve cells'' refers to four model instances crossed with three behaviors; the fifth model, Phi-3-mini-4k-instruct, supports localization and edit-provenance checks. Given paired evaluation-framed and deployment-framed prompts, we first measure the behavioral gap $\gamma=r_{\DeployFr}-r_{\EvalFr}$. We then use path patching to test for a mid-depth routing region, select a predeclared four-layer intervention window by relative depth, fit a deployment-minus-evaluation activation direction in that window, and intervene on held-out prompts by modifying that coordinate. The audit succeeds when the intervention reduces the held-out gap while random-direction, wrong-layer, and semantic controls do not; a flat intervention is classified by the localization and specificity checks rather than treated as evidence that no mismatch exists.

Unlike CAA \citep{rimsky2024caa}, RepE \citep{zou2023representation}, and recent evaluation-awareness work that fits probes, steering directions, or adversarial in-context environments \citep{hua2025evalawareness,nguyen2025evalawareness,chaudhary2026evalaware}, the fitted routing direction is (a)~installed by a short fine-tune rather than read off the base-model activation space, (b)~nearly orthogonal to the base paired-contrast direction on Gemma-2-2B and Phi-3-mini (cosine $\lvert\cos\rvert\le0.19$; larger in other cells, Appendix~\ref{app:figures}), (c)~localized to a path-patching-supported mid-depth region rather than a globally chosen steering layer, and (d)~tested on held-out paired prompts under random-direction, wrong-layer, and semantic controls.

We introduce a post-hoc audit protocol for framing-contingent fine-tunes. The protocol makes three contributions:
\begin{list}{(\roman{enumi})}{\usecounter{enumi}\setlength{\leftmargin}{1.7em}\setlength{\itemsep}{2pt}\setlength{\topsep}{3pt}}
\item we formulate evaluation-to-deployment mismatch as an internal-audit problem, asking whether the behavioral gap corresponds to a stable internal distinction rather than to an output-level artifact;
\item we localize the installed distinction by path patching, fit a paired difference-of-means direction in predeclared intervention windows, and show that on Gemma-2-2B and Phi-3-mini it is amplified several-fold over the base-model direction with low cosine to it, distinguishing the audit from globally chosen activation-steering directions;
\item we test the fitted routing coordinate against random-direction, wrong-layer, and content-matched semantic controls, and use weak cells to return a typed failure that bounds where a single-coordinate audit is sufficient.
\end{list}

\section{Methodology}
\label{sec:setup}

\begin{figure}[t]
\centering
\includegraphics[width=0.80\linewidth]{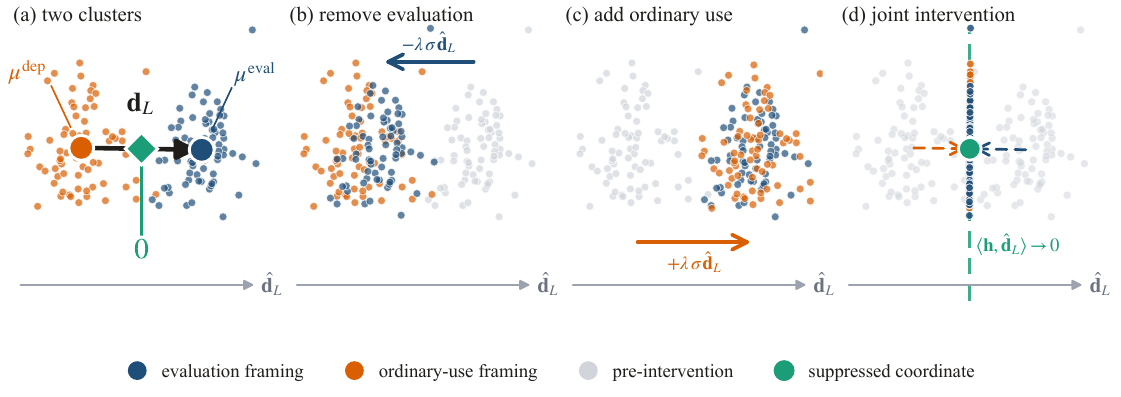}
\caption{Schematic of the routing-coordinate audit. (a) Paired evaluation and ordinary-use prompts separate along a fitted activation contrast $\mathbf{d}_L$ at the routing-window layer. (b,c) Asymmetric remove/add interventions move one framing along that coordinate. (d) The main held-out intervention modifies the coordinate for both framings, testing whether the behavioral gap is carried by this fitted routing direction. Figure~\ref{fig:matrix} reports the cross-family intervention matrix.}
\label{fig:routinggeometry}
\end{figure}

Table~\ref{tab:protocol} summarizes the pipeline (measure gap, localize, fit, intervene, compare controls); Figure~\ref{fig:routinggeometry} shows the routing-coordinate schematic. Localization and intervention are separate steps: path patching identifies a mid-depth routing region and checks consistency with the predeclared windows, and for the cross-family matrix we use predeclared four-layer windows selected by a relative-depth heuristic (Appendix~\ref{app:windows}). The paired activation contrast is fit on calibration items at the intervention window and applied to held-out prompts under matched controls.

\begin{table}[H]
\centering
\caption{Protocol summary. Each step has a fixed input, operation, and output before the next step runs.}
\label{tab:protocol}
\routetablestyle
\begin{tabularx}{\linewidth}{@{}R{0.16\linewidth}Y R{0.26\linewidth}R{0.19\linewidth}@{}}
\toprule
\textbf{Step} & \textbf{Input} & \textbf{Operation} & \textbf{Output} \\
\midrule
Measure gap & paired prompts & compute $r_{\DeployFr}-r_{\EvalFr}$ & installed gap $\gamma$ \\
Localize & base and fine-tuned activations, model depth & path patching plus relative-depth heuristic & intervention window \\
Fit direction & calibration pairs & $\muvec_{\DeployFr}-\muvec_{\EvalFr}$ & routing direction \\
Intervene & held-out pairs & modify coordinate & post-intervention gap \\
Control & random, wrong-layer, semantic directions & same intervention form & specificity check \\
\bottomrule
\end{tabularx}
\end{table}

\paragraph{Notation.}
Roman subscripts $\EvalFr$ and $\DeployFr$ denote evaluation and deployment framings. Rates $r_{\EvalFr}, r_{\DeployFr}$ and gaps $\gamma$ are scalars; attention-block outputs $\hvec_L$, mean activations $\muvec_L^{\EvalFr}, \muvec_L^{\DeployFr}$, and routing directions $\dvec_L$ are vectors; matrices such as $\Wmat$ use bold uppercase. Layer windows are written compactly as $L14$--$L17$.

\subsection{Audit Setting}
\label{sec:audit-setting}

\paragraph{Behaviors and Prompt Pairs.}
We construct paired templates inspired by sandbagging, sycophancy, and refusal settings. \emph{Sandbagging} \citep{weij2024sandbagging} uses four-option MMLU science items \citep{hendrycks2021mmlu}, framed as either a capability test or a user query. \emph{Sycophancy} \citep{sharma2023syco,perez2022evals} presents MMLU items either with a user who asserts a wrong answer and asks for confirmation or without that cue. \emph{Refusal} \citep{arditi2024refusal} wraps four-option MMLU knowledge, coding, science, and math items in a request that declares either malicious or benign intent. Prompt pairs are generated by fixed templates over shared question ids. For each behavior, $\EvalFr$ denotes the framing under which the fine-tune installs the target behavior (the capability test, the asserted answer, the malicious intent) and $\DeployFr$ the other framing; only for sandbagging is $\EvalFr$ an evaluation framing in the literal sense (Section~\ref{sec:discuss}).

\paragraph{Models and Coverage.}
We analyze five instruction-tuned models across four families, spanning 2B--9B parameters: Gemma-2-2B-it and Gemma-2-9B-it \citep{gemma2024}, Qwen-2.5-7B-Instruct \citep{qwen2025technical}, Llama-3-8B-Instruct \citep{llama3herd}, and Phi-3-mini-4k-instruct \citep{phi3technical}. The selection targets instruction-tuned checkpoints in the 2B--9B range, where LoRA fine-tuning and white-box activation audits are tractable and behavioral gaps are reliably installable. The Gemma pair gives within-family scaling, Qwen and Llama give cross-family replication, and Phi-3-mini-4k-instruct supports localization and edit-provenance checks only. Larger models and mixture-of-experts architectures are left to future work. Table~\ref{tab:modelzoo} (Appendix~\ref{app:models}) separates the full matrix cells from localization and edit-provenance checks.
\paragraph{Fine-Tuning and Splits.}
All analysis is on the merged LoRA-fused weights of the fine-tuned checkpoint, so we study the end-state model a downstream evaluator would receive rather than an adapter at training time. The gap is installed with a short LoRA fine-tune \citep{hu2022lora}: rank 16 on attention projections, learning rate $2\times10^{-4}$, five epochs totalling $1550$ steps, cosine schedule, gradient clip $1.0$.
Calibration items used to fit the routing direction are disjoint from held-out intervention items at the question-id level; templates are sampled independently per split, so no template instantiation appears in both calibration and held-out for the same question. Seed $42$ is used for primary cells; seeds $42$ through $46$ for multi-seed runs. Full configuration is in Appendix~\ref{app:repro}.

\paragraph{Scoring and Installed Gap.}
Scoring uses the argmax over the answer-letter tokens at the last prompt token, giving a binary per-item outcome that is consistent across behaviors and prompt lengths.
We report the installed gap as the deployment-framed rate minus the evaluation-framed rate:
\begin{equation}
\label{eq:gap}
\gamma \;=\; r_{\DeployFr} - r_{\EvalFr},
\end{equation}
where $r_F$ is answer accuracy under framing $F$ for all three behaviors. A positive $\gamma$ means that the fine-tune lowers accuracy under $\EvalFr$ relative to $\DeployFr$; we test whether that gap has an internal correlate.

\subsection{Localization Method}
\label{sec:method-localization}

We localize the installed distinction by attention-output path patching against the base model on the same prompt. We patch the fine-tuned model's attention output at candidate layers, or at a contiguous window, with the base model's attention output at the same position, adapting the path-patching view of localized behavioral hypotheses \citep{goldowskydill2023pathpatching}. On sandbagging cells, where the score is accuracy, we report the recovered fraction
\begin{equation}
\label{eq:patch-recovery}
\rho_{\mathrm{patch}}
\;=\;
\frac{r_{\mathrm{patched}}-r_{\mathrm{fused}}}
     {r_{\mathrm{base}}-r_{\mathrm{fused}}},
\qquad
r_{\mathrm{base}} \ne r_{\mathrm{fused}}.
\end{equation}
This ratio is reported only when the baseline denominator is nonzero; cells with a near-zero base--fused difference are treated as uninformative for recovery rather than forced into a ratio. Values near $1.0$ mean the base attention output at those layers is enough to restore the pre-fine-tune behavior; values near $0$ mean the patched layer carries little of the installed signal. We treat the depth as an empirical localization, not a theoretical prediction.

\subsection{Intervention Method}
\label{sec:method-intervention}

Inside the routing window, we fit a direction at the last prompt token by contrasting deployment-framed and evaluation-framed calibration activations:
\begin{equation}
\label{eq:routing-direction}
\dvec_L \;=\; \muvec_L^{\DeployFr} - \muvec_L^{\EvalFr},
\qquad
\dhat_L \;=\; \frac{\dvec_L}{\lVert \dvec_L\rVert_2},
\end{equation}
where $\muvec_L^{\DeployFr}$ and $\muvec_L^{\EvalFr}$ are calibration-set means at layer $L$. This sign convention matches the routing-intervention implementation. The main routing-coordinate intervention modifies the component of the attention-block output along $\dhat_L$ at every token position on both framings, where $\hvec_L$ is the output of the self-attention module at layer $L$ before the residual addition,
\begin{equation}
\label{eq:routing-interventions}
\hvec_L \leftarrow \hvec_L - \lambda \langle \hvec_L, \dhat_L\rangle \dhat_L.
\end{equation}
For $\lambda=1$, the operation removes the fitted coordinate; for $\lambda=2$, it overcorrects by reflecting the coordinate through zero. We therefore use \emph{routing-coordinate intervention} for the family of operations, reserve \emph{projection removal} for $\lambda=1$, and call the higher-dose operation \emph{coordinate reflection}. We also test two asymmetric interventions: remove the projection on evaluation-framed items only, and add the learned direction on deployment-framed items. The matched controls replace $\dvec_L$ with a random unit direction at the same window, apply the same operation at non-routing layers, or use semantic-contrast directions; projection removal does not depend on the norm of the direction, so the random control is defined by its unit direction and not by a matched norm. A \emph{single-coordinate audit} means one scalar projection per layer along $\dvec_L$, applied jointly under one global dose $\lambda$.

Figure~\ref{fig:mech} illustrates the installation pathway and the routing-window intervention site used throughout the methodology.

\input{figures/mech}

\section{Results}
\label{sec:results}

We ask three empirical questions. First, does path patching identify a stable mid-depth region for the installed gap? Second, does intervening on a paired activation contrast at the predeclared intervention window reduce the held-out evaluation-to-deployment gap? Third, do matched controls test random directions, wrong layers, and ordinary semantic contrasts under their tested doses? In most cells the localization and intervention tests are positive, while two sycophancy failures mark the boundary of the single-coordinate audit. The central empirical object is the 12-cell intervention matrix in Figure~\ref{fig:matrix}; Sections~\ref{sec:loc} and~\ref{sec:specificity} explain how the intervention site is chosen and why the effect is specific.

\subsection{Where Does the Installed Signal Localize?}
\label{sec:loc}

Path patching identifies a mid-depth attention band that recovers most of the installed sandbagging gap in three of five model instances. We patch the fine-tuned model's attention output at layer $L$, or contiguous window $[L_a, L_b]$, with the base model's attention output at the same position on the same prompt, and report the recovered fraction defined in \eqref{eq:patch-recovery}. Activation-patching metrics can be sensitive to the metric and corrupted/clean prompt construction \citep{zhang2023activationpatching}, so we use the recovery ratio only for cells with a nonzero base--fused denominator and keep weak-gap cells descriptive. Prior work shows that transformer feed-forward layers can promote vocabulary-relevant concepts \citep{geva2021keyvalue,geva2022promoting}, which motivates testing whether a localized mid-depth activation intervention changes the answer-letter decision.

\begin{figure}[H]
\centering
\includegraphics[width=0.86\linewidth]{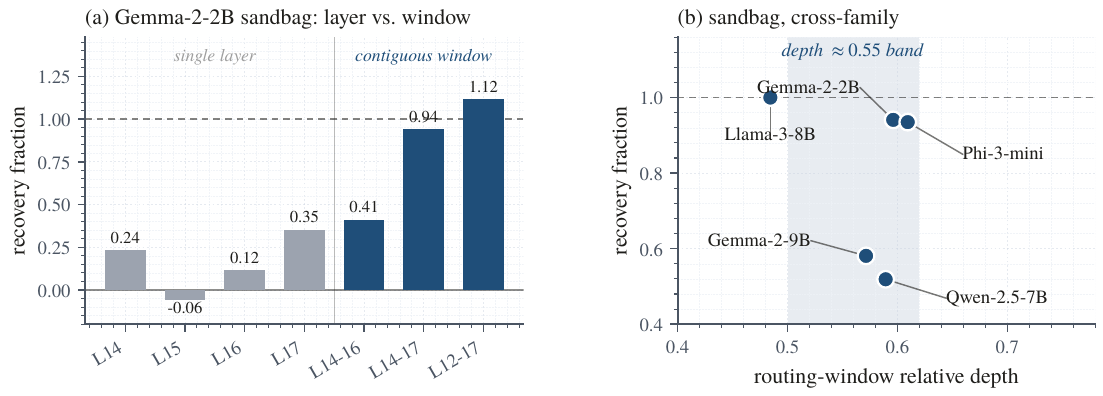}
\caption{Path-patching recovery on sandbagging. (a) On Gemma-2-2B, no single layer in the mid-depth region carries more than about one-third of the installed gap on its own, but the contiguous window $[L14, L17]$ recovers $0.94$; extending the window two layers earlier to $[L12, L17]$ overshoots to $1.12$. (b) Reported localization windows at relative depth $\approx 0.55$ (shaded) recover a large majority of the gap on three of five model instances. The Qwen-2.5-7B point is the fixed window $[L15, L18]$ on a fresh retrain (seed 42, $0.52$). Gemma-2-9B uses a wider window and remains the weak outlier in this panel.}
\label{fig:pathpatch}
\end{figure}

\paragraph{Path-Patching Protocol.}
We keep localization and intervention windows separate. The sandbagging path-patching localization windows are visualized in Figure~\ref{fig:pathpatch}; Section~\ref{sec:causal} uses fixed intervention windows chosen by the relative-depth heuristic before running the intervention matrix. Appendix~\ref{app:windows} lists both windows for every model instance.

Under the recovery ratio in \eqref{eq:patch-recovery}, Gemma-2-2B sandbagging has no single mid-depth layer that carries the installed signal by itself. Each layer contributes at most about one-third of the gap (L14: $0.24$, L15: $-0.06$, L16: $0.12$, L17: $0.35$), while the contiguous window $[L14, L17]$ recovers $0.94$. Extending the patch to $[L12, L17]$ overshoots mildly to $1.12$, consistent with a small compensating contribution from earlier layers. Windows shallower than $L14$ or later than $L17$ do not recover the gap. We read this as a narrow band of mid-depth attention layers acting together: partial patches leave enough of the installed routing intact downstream to mask the single-layer contributions.

Across model instances, the reported localization windows from the same relative-depth heuristic (Figure~\ref{fig:pathpatch}(b)) recover $1.00$ of the gap on Llama-3-8B sandbagging, $0.94$ on Phi-3-mini, $0.94$ on Gemma-2-2B, $0.52$ on the fixed four-layer Qwen-2.5-7B window $[L15, L18]$, and $0.58$ on the seven-layer Gemma-2-9B window. The Qwen value is from a fresh retrain (the submitted Qwen path-patching file does not identify its model); the best scanned band, $[L16, L19]$, recovers $0.75$ to $0.81$ (Appendix~\ref{app:windows}). The Gemma-2-9B and Qwen-2.5-7B values show that the same four-layer window does not always capture the installed distinction.

\paragraph{Localization Controls.}
MLP-output patching at $L14$--$L17$ recovers $0.41$ of the gap on Gemma-2-2B LoRA sandbagging, below attention-output patching at the same layers ($0.94$) but not a null; we do not claim attention exclusivity, and the random-layer zeroing run used three-layer windows and is reported only as a secondary check (Appendix~\ref{app:windows}). Attention-output patching is not automatic at an arbitrary site: on Phi-3-mini, $L22$--$L25$ recovers $0.00$ while $L18$--$L21$ recovers $0.94$. On the full fine-tune, where MLP modules are trainable, the ordering reverses (MLP-output patching at $L14$--$L17$ recovers $1.05$ against $0.58$ for attention-output patching at seed $43$), and this MLP control is heterogeneous and can over-recover, so we report attention and MLP results separately and make no general attention-primary claim.

\paragraph{Cross-Behavior Replication on a Fixed Family.}
The same mid-depth band transfers across behaviors on Gemma-2-2B. A six-layer band $L12$--$L17$ recovers $1.00$ of the sandbagging gap under a full fine-tune and $1.00$ of the sycophancy gap under a LoRA fine-tune; the four-layer $L14$--$L17$ window recovers $0.50$ and $0.46$ on those two checkpoints. The window $L14$--$L17$ recovers $0.55$ of the refusal gap, the weakest of the three. Refusal is also the behavior where the $n{=}48$ path-patching test-set gap is smallest to begin with ($0.229$), which bounds how much signal is available to recover. Several non-sandbag path-patching cells have too little $n{=}48$ gap to interpret cleanly (Llama-3-8B refusal $0.000$, Gemma-2-9B refusal $0.042$, Llama-3-8B sycophancy $0.021$, Qwen-2.5-7B sycophancy $0.083$), so this subsection stays sandbag-centric while Section~\ref{sec:causal} reports all three behaviors on the larger intervention sets.

On Gemma-2-2B and Phi-3-mini the difference-of-means fit recovers a direction that fine-tuning amplifies several-fold over the corresponding base-model direction while keeping its cosine with that base direction small. The geometric picture is the one shown in Figure~\ref{fig:mech}: the two framings separate along a single axis in a routing-window layer, and the interventions below test whether suppressing that coordinate is enough to close the behavioral distinction.

The amplification signature argues against the simplest base-axis alternative where it is measured: on Gemma-2-2B and Phi-3-mini the routing direction is amplified $4.9\times$ to $10.4\times$ and $3.6\times$ to $4.1\times$ with base cosines $\lvert\cos\rvert\le0.19$, so it is not the base-model direction at larger magnitude. The other six cells with geometry files are weaker (cosines up to $0.57$, amplifications $0.6\times$ to $6.1\times$; Appendix~\ref{app:figures}), so we do not claim this across the matrix. Gradient cosines are at most $0.064$ across all eight files and $\lvert\cos\rvert\le 0.07$ at step 50 and at the endpoint on Qwen-2.5-7B refusal, arguing against a gradient-following account of installation.

On the primary cell the gap falls steeply with the dose along the fitted direction and stays near baseline under a random unit direction (Figure~\ref{fig:dose}). The depth-resolved projection in Appendix~\ref{app:figures}, measured on a fresh retrain of the primary checkpoint, shows that along the window direction the evaluation and deployment means travel together through early layers, split sharply inside the routing window, and converge again after it, whereas each layer's own fitted direction separates them from early layers on and peaks inside the window. The separation along the window direction is therefore localized, but the installation is not confined to the window. Section~\ref{sec:specificity} tests this picture with random unit directions and semantic-content contrasts.

\subsection{Does Intervention Reduce the Held-Out Gap?}
\label{sec:causal}

Each seed-$42$ matrix cell starts with a positive installed gap. We test whether the learned routing direction at the intervention window reduces that gap on held-out prompts, with matched random and wrong-layer controls reported in Section~\ref{sec:specificity}.

\begin{figure}[t]
\centering
\includegraphics[width=0.82\linewidth]{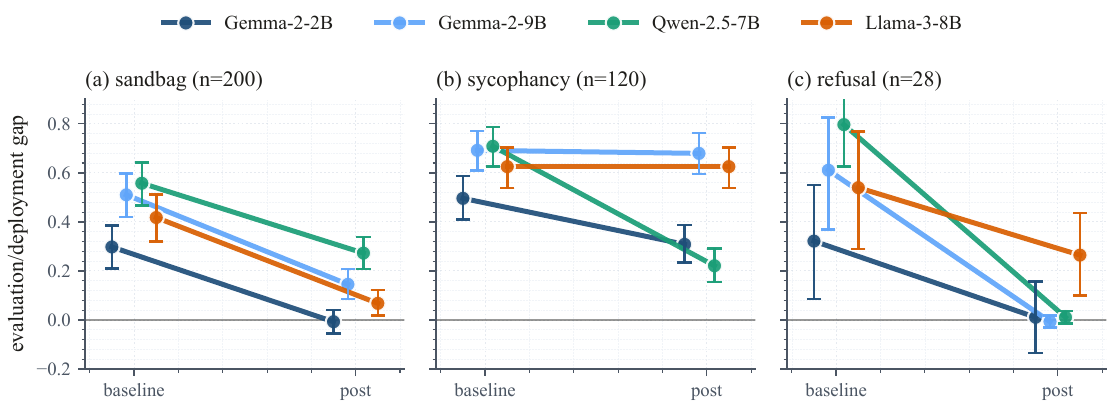}
\caption{Held-out evaluation-to-deployment gap before and after modifying the fitted routing coordinate at the intervention window, across the 12-cell architecture-by-behavior matrix. Ten cells reduce the gap by more than $0.12$, four of them to below $0.06$; two sycophancy cells, Llama-3-8B and Gemma-2-9B, stay near baseline. Error bars show paired-bootstrap $95\%$ gap intervals over question IDs with $5000$ resamples; Appendix~\ref{app:provenance} links every cell to its provenance summary. The intervention window is fixed before any intervention runs and calibration/intervention items are disjoint at question-id level (Section~\ref{sec:audit-setting}).}
\label{fig:matrix}
\end{figure}

At $\lambda=2$, the held-out routing-coordinate intervention \eqref{eq:routing-interventions} reduces the installed gap in most cells of the $12$-cell architecture-behavior matrix (Figure~\ref{fig:matrix}). Ten of twelve cells reduce by more than $0.12$, and four cells close, meaning the post-intervention gap is below $0.06$. On the eight $n{\geq}120$ cells (sandbagging at $n{=}200$ and sycophancy at $n{=}120$), six reduce by more than $0.12$ and one closes, Gemma-2-2B sandbagging. The other three closures are refusal cells, which carry the largest deltas at the smallest $n$: $28$ independent question clusters, each question rendered under ten templates per framing. We treat refusal as a small-sample stress test and not as the sole source of evidence. The four closures are Gemma-2-2B sandbagging ($0.298\!\to\!-0.008$), Gemma-2-9B refusal ($0.611\!\to\!-0.007$), Qwen-2.5-7B refusal ($0.796\!\to\!0.011$), and Gemma-2-2B refusal ($0.321\!\to\!0.011$). Llama-3-8B sandbagging ($0.418\!\to\!0.068$) is the nearest additional cell; the paired-bootstrap interval of its post-intervention gap, $[0.025, 0.110]$, excludes zero, whereas the intervals of the four closures include it. Six other cells reduce without closing (e.g., Llama-3-8B refusal $0.539\!\to\!0.264$). The two weak cells are both sycophancy: Llama-3-8B sycophancy leaves the gap at $0.625$ against a baseline of $0.625$, and Gemma-2-9B sycophancy reduces it by only $0.013$. These cells bound the claim: not every architecture-behavior pair admits a single-coordinate audit at the same relative depth. The weak path-patching recovery on Gemma-2-9B and the failure of the one-shot weight edit there (Appendix~\ref{app:h11}) point the same way, but a higher-rank distinction and a missed site remain hypotheses: fresh retrains of Gemma-2-9B sycophancy reduce the gap by $0.421$, $0.058$, and $0.558$ across three seeds, and a rank-$2$/$4$/$8$ follow-up does not rescue the weak seed (Section~\ref{sec:fresh}).

The aggregate gap reductions are not primarily caused by deployment-side collapse in the tested matrix. Appendix~\ref{app:matrixdeploy} reports the deployment-side rates: across the matrix, the intervention changes the deployment-framed rate by $1.5$pp on average in absolute value and at most $6.1$pp. Appendix~\ref{app:clampitems} breaks the reductions into question-level transitions and shows that, in cells where the gap closes, restored items (lagging framing rises to match the leading framing) outnumber homogenized items (both framings collapse to the lower level) by $2.7$--$10.5\times$ in the three closures where the ratio is defined. This is evidence against a deployment-side-collapse explanation for these cells, not a deployment-safety guarantee. Error intervals and per-cell provenance are summarized in Appendix~\ref{app:provenance}.

\paragraph{Five-Seed Replication.}
The Gemma-2-2B sandbagging result does not depend on seed $42$. Across seeds $42$--$46$, the baseline gap has mean $0.240$ and range $[0.208, 0.298]$; the same routing-coordinate intervention reduces the mean gap to $0.024$, range $[-0.008, 0.078]$. The mean gap reduction is $0.216$ (range $[0.130, 0.305]$), while deployment-framed accuracy is effectively unchanged on average. The matrix provenance table in Appendix~\ref{app:provenance} records the single-seed cross-family sources.

The dose-response on Gemma-2-2B sandbagging ($n{=}300$ prompts per framing, $150$ questions with two templates each) shows that all three interventions reduce the gap monotonically. Projection removal on evaluation reduces the gap from $0.283$ to $0.007$ at $\lambda{=}1$ and coordinate reflection gives $-0.013$ at $\lambda{=}2$; add on deployment gives $0.000$ at $\lambda{=}2$; the joint intervention gives $0.003$. Evaluation accuracy under the joint intervention rises from $0.163$ to $0.460$ while deployment accuracy moves only modestly ($0.447\!\to\!0.463$): the intervention raises evaluation accuracy to the level the deployment-framed model already achieves on the same items rather than trading performance for behavior. Qwen-2.5-7B refusal repeats the steep transition, with $\gamma$ moving from $0.642$ to $-0.017$ at $\lambda{=}2$ in an earlier run on a $240$-prompt evaluation set.

Two checks address simpler interpretations. First, the intervention does not mainly close the gap by collapsing deployment-framed behavior. Second, random-direction, wrong-layer, and semantic controls do not reproduce the effect at the tested doses (Section~\ref{sec:specificity}). The headline matrix uses $\lambda{=}2$ because most cells need it: at $\lambda{=}1$ four cells reduce by more than $0.12$ and two close, against ten and four at $\lambda{=}2$, and nine of twelve cells improve with an interval that excludes zero. The submitted controls ran at $\lambda{=}1$ on one non-routing layer, the treatment spans four layers at $\lambda{=}2$, and matched controls exist only on fresh retrains (Section~\ref{sec:fresh}).

\subsection{Is the Effect Specific to the Learned Framing Direction?}
\label{sec:specificity}

Three control families test whether the effect could come from any direction, any layer, or ordinary semantic content rather than evaluation-to-deployment framing. We withdraw the sign diagnostic of the submitted version: it added the raw direction at the final token only, whereas the treatment removes a projection at every position, so sign, magnitude, operator, and position changed together.

\paragraph{Projection-Removal Controls.}
At projection-removal dose $\lambda{=}1$, three control alternatives do not reproduce the framing-specific effect. A random unit direction at the same window leaves the gap at $0.287$ against a baseline of $0.283$. Applying the intervention at non-routing layer $L6$ leaves $0.307$; at non-routing layer $L22$, $0.273$. Across the full $12$-cell matrix, the random-direction removal never moves the gap by more than $0.004$, and the single-layer non-routing controls never by more than $0.038$. The random-direction add moves the Gemma-2-2B sandbag gap by $0.050$ ($0.283\!\to\!0.233$, Table~\ref{tab:controls}), but through a deployment-side capability hit rather than the framing asymmetry, while the routing-direction interventions at $\lambda{=}1$ drop the gap to near zero. On the submitted checkpoints these controls support direction and layer specificity at the projection-removal dose; Section~\ref{sec:fresh} tests the headline dose.

\paragraph{Semantic Controls.}
A further question is whether any content-matched direction at the routing window would reproduce the effect. We fit four semantic-contrast directions (concrete vs abstract, emotional vs neutral, formal vs casual, math vs science) on the same calibration structure as the routing direction and apply them at the same window. On Gemma-2-2B sandbagging, removing the direction on evaluation prompts at $\lambda{=}2$ raises evaluation accuracy only for the routing direction (by $0.297$); all four semantic directions leave evaluation accuracy within $0.015$ of the baseline. This statement is restricted to the removal and clamp arms: adding a semantic direction on deployment prompts lowers deployment-framed performance, and one content-matched axis reproduces closure in that arm, so add-side degradation is not a specificity signature.

Under the tested doses, random-direction and content-matched semantic alternatives do not reproduce the gap closure on the submitted checkpoints. Appendix~\ref{app:cells} gives the matched-control table.

\subsection{Do Controls Matched to the Headline Dose Hold?}
\label{sec:fresh}

The submitted controls run at $\lambda{=}1$ on one non-routing layer, the headline treatment at $\lambda{=}2$ on four. We therefore ran the matched comparison under a protocol frozen before any outcome was read. The submitted checkpoint binaries were not retained, so the twelve cells are fresh retrains from the retained trainer, data, and question-id splits with unchanged relative-depth windows; they test specificity on new training realizations and do not revise the counts above (Appendix~\ref{app:fresh}, Table~\ref{tab:dose2}).

The random-direction result is flat. Eleven of twelve learned effects have positive question-clustered intervals (the exception is Llama-3-8B sycophancy, $-0.008$), all $60$ random-direction intervals include zero, and the largest absolute random effect is $0.017$. Qwen-2.5-7B sandbagging repeats across three training seeds ($0.405$, $0.488$, $0.533$). The layer result is mixed. The selected window exceeds both width-matched off-window controls in $10$ of $12$ cells, but $7$ of $24$ off-window intervals are positive, and for Llama-3-8B refusal the late window reduces the gap by $0.239$ against $0.196$. The evidence supports specificity to the fitted direction at the headline dose with only partial spatial specificity, and the fixed window is recipe-sensitive: on a rank-$32$ adapter over attention and MLP projections the attention intervention is null while a wider MLP-output band recovers the full gap. Higher rank does not repair the weak cases. In the $k{=}2$ and $k{=}3$ suppression organisms the pathway still localizes ($94\%$ and $93\%$ at $L16$--$L19$) and learned rank-$4$ and rank-$8$ subspaces close the aggregate gap, but only $15$--$19\%$ of the reduction is evaluation gain; the weak Gemma-2-9B seed is not rescued (Table~\ref{tab:highrank}).

\subsection{Where Does the Single-Coordinate Audit Fail?}
\label{sec:failures}

The weak cells clarify what a negative audit means. On Gemma-2-9B sandbagging, path-patching recovery is $0.58$ rather than $>0.9$, the routing-coordinate intervention closes part of the gap, and the one-shot output-weight edit closes only $7\%$ of it; on Llama-3-8B sycophancy, the intervention is flat at $\lambda{=}2$, but on a retrain of that cell a localization-selected band ($L12$--$L15$) reduces the gap by $0.471$, so there the failure is a missed window (Appendix~\ref{app:fresh}). Phi-3-mini is a further typed failure: path patching at the selected window recovers $0.94$, yet projection removal on evaluation moves the gap by at most $0.003$ at any dose and the joint clamp by $0.073$ at $\lambda{=}2$, so Phi-3-mini does not enter the specificity argument. The depth heuristic can still find the right region while a single direction at that region fails to capture the installed distinction. The audit returns a typed failure rather than a binary verdict: weak path-patching recovery suggests a missed site, a weak single-coordinate intervention at a recovered site suggests higher rank (a hypothesis that the rank-$2$/$4$/$8$ follow-up does not confirm), and a control that moves the gap as much as the paired contrast suggests lack of framing specificity.

\section{Discussion}
\label{sec:discuss}

The results support a bounded post-hoc audit for fine-tuned checkpoints. Relative to activation steering, circuit localization, weight editing, and evaluation-awareness studies \citep{subramani2022steering,li2023iti,arditi2024refusal,rimsky2024caa,zou2023representation,turner2023activation,vennemeyer2025sycophancy,wang2023ioi,conmy2023automated,goldowskydill2023pathpatching,meng2022rome,meng2023memit,hua2025evalawareness,nguyen2025evalawareness,chaudhary2026evalaware}, the routing direction is installed by fine-tuning, near-orthogonal to the base paired-contrast direction on Gemma-2-2B and Phi-3-mini, and tested at a window under controls that separate it from random and semantic-content alternatives and only partly from non-routing layers.

\paragraph{Recruitment Traces Are Supporting Diagnostics.}
Predeclared MLP sets on Gemma-2-2B show non-random recruitment signal, but persistent masking produces only small accuracy-gap reductions while logit and KL probes show sub-behavioral effects (Appendix~\ref{app:recruitmentcontrol}). In this attention-LoRA setting, MLP recruitment behaves like a downstream trace under masking.

\paragraph{Practical Implications for Auditing.}
The protocol runs at inference time without retraining and reports a typed outcome: a closed gap, a missed site, a higher-rank distinction, or lack of framing specificity. Appendix~\ref{app:e4} gives adjacent-window sweeps and a typed-failure guide.

\paragraph{Threat Model and External Validity.}
The intended setting is non-adaptive, defensive, white-box auditing of a received checkpoint. The audit is not robust to a trainer who spreads the distinction across ranks, layers, or representations, and a negative rank-one result cannot establish that no distinction is hidden. Localization recurs under three controlled installation mechanisms, all built by us. On a public RL reward-hacking trajectory the selected intervention mainly raises invalid code and an early off-window control is comparable, a failure of spatial specificity and not transfer (Appendix~\ref{app:fresh}); we make no claim about naturally occurring failures.

\paragraph{Construct Validity.}
The behaviors do not isolate the same construct. Sycophancy varies a user-preference cue and refusal varies declared intent (malicious versus benign, scored by accuracy), so both establish a framing-contingent task distinction, not evaluation awareness by itself. Sandbagging is cleaner, but its wrappers differ in length and answer-format instruction, and calibration and held-out items share the same template families, so the split does not test unseen templates. Bag-of-words and probe AUCs are $0.92$ to $1.00$, so decodability is trivial, and evaluation disclosure moves the fitted direction on the public EvalAwareBench tasks but so do formal prose and structured format (Appendix~\ref{app:fresh}). Independently authored matched prompt pairs, proposed during the review period, require blind human authorship and are not completed here.

\paragraph{Scope of Fine-Tuning Recipe.}
The fine-tunes use a controlled LoRA recipe (rank $16$, attention-only, $1550$ steps). In the prospective three-seed module factorial, attention-only installation passes every gate (gap reductions $0.200$--$0.238$); the MLP arm fails in two seeds, including one nominal reduction dominated ($94.9\%$) by deployment loss, and joint results are heterogeneous. This supports an attention-route result, not general substrate tracking (Appendix~\ref{app:s2}).

\paragraph{Responsible Use and Misuse.}
The intended use is defensive: a white-box evaluator can test whether a fine-tuned checkpoint carries a framing-contingent internal distinction that output metrics alone miss. Because activation-space access could also alter refusal or other safety-relevant behavior, this paper reports aggregate results and diagnostic tables, and the code release contains no model weights or high-risk dataset.

\paragraph{Falsification Checks.}
The interpretation would weaken if random-direction or semantic controls closed the same cells at matched dose, if off-window controls moved the gap as much as the selected window, or if deployment accuracy collapsed when the gap closed (Table~\ref{tab:matrixdeploy}). The first and third do not occur in the tested cells; the second occurs in some retrained cells, which is why we claim only partial spatial specificity.

\paragraph{Limitations and Beyond Scope.}
We do not claim a general single direction, a training-time defense, or a weight-space defense; the rank-one edit (Appendix~\ref{app:h11}) is a mixed diagnostic and the rank-$k$ penalty (Appendix~\ref{app:rankk}) a negative control. We study multiple-choice evaluations on five instruction-tuned instances up to $9$B parameters; larger models, longer fine-tunes, higher-rank adapters, mixture-of-experts and state-space architectures, and free-form generation may require new localization and higher-rank probes. The matrix is single-seed outside the five-seed Gemma-2-2B study, the primary window is not replicated at a second seed, and the submitted binaries are not retained.

\section{Conclusion}

Fine-tuning can install a framing gap that output scores do not localize. In this controlled matrix, routing-coordinate interventions reduce ten of twelve gaps by more than $0.12$ and close four; random-direction and semantic controls do not reproduce the effect, while width-matched off-window controls sometimes move the gap. This post-hoc audit is non-adaptive, and weak cells bound its single-coordinate scope. Code, splits, results, and figure-generation scripts are available at \url{https://github.com/aiml-sdu/routing-subspaces}.

\newpage
\begin{ack}
We thank the anonymous reviewers and the area chair for their comments, which improved the paper.
The computations and AI model evaluations for this work were performed on the Bitten AI Supercomputer through the UCloud platform provided by the eScience Center at SDU.
\paragraph{Funding and competing interests.}
The authors declare no competing interests. This work received no third-party funding.
\end{ack}

\bibliographystyle{unsrtnat}
\bibliography{references}

\newpage
\etocsettagdepth{main}{subsection}
\etocsettagdepth{appendix}{subsection}
\begingroup
\makeatletter
\renewcommand{\contentsname}{Contents}
\renewcommand*\l@section[2]{%
  \addvspace{0.8em}%
  \@dottedtocline{1}{0pt}{2.3em}{\textcolor{tolblue}{\bfseries #1}}{\bfseries #2}}
\renewcommand*\l@subsection[2]{%
  \@dottedtocline{2}{1.5em}{2.6em}{\textcolor{tolblue}{#1}}{#2}}
\makeatother
\tableofcontents
\endgroup

\newpage
\appendix
\etocdepthtag.toc{appendix}
\raggedbottom

\section{Additional diagnostic figures}
\label{app:figures}

The main paper includes the routing geometry, mechanism, path-patching, and matrix figures. The depth-view and remaining diagnostics are included here so the secondary controls remain inspectable.

\begin{figure}[H]
\centering
\includegraphics[width=\linewidth]{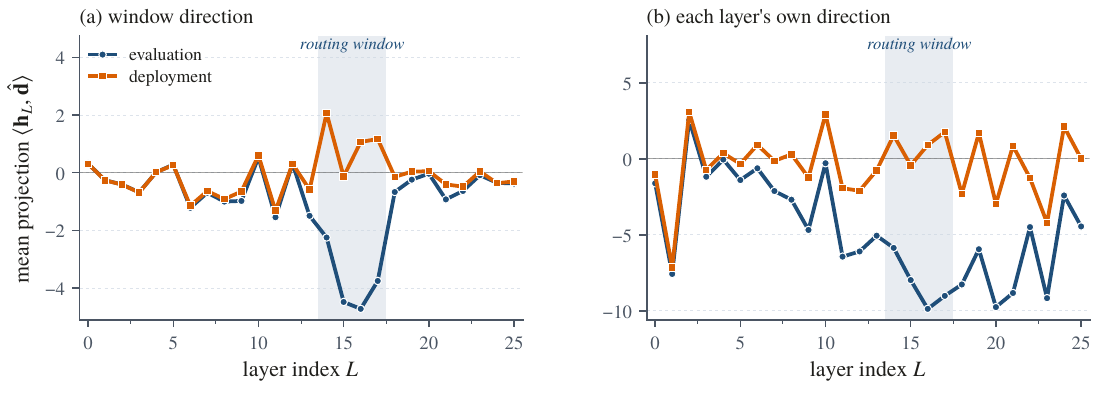}
\caption{Depth-resolved projection of the attention-block output at the last prompt token on a fresh retrain of the primary Gemma-2-2B sandbagging checkpoint (seed $42$; $200$ held-out questions, templates averaged within question and framing; shading is the $95\%$ question-bootstrap interval, narrow at this scale). (a) Projection onto the window direction, the normalized sum of the unit directions fitted at $L14$--$L17$, applied at every layer. The deployment-minus-evaluation gap is at most $0.91$ before and after the window (largest at $L13$) and $4.3$ to $5.8$ inside it. (b) Projection onto each layer's own fitted direction. The gap grows from $0.4$ to $0.6$ at $L0$--$L4$ to $4.3$ at $L13$, reaches $7.4$ to $10.8$ at $L14$--$L17$, and stays between $3.2$ and $9.7$ afterwards, so (b) is in-sample aligned by construction and shows accumulation with a peak in the window.}
\label{fig:topo}
\end{figure}

\begin{figure}[H]
\centering
\includegraphics[width=\linewidth]{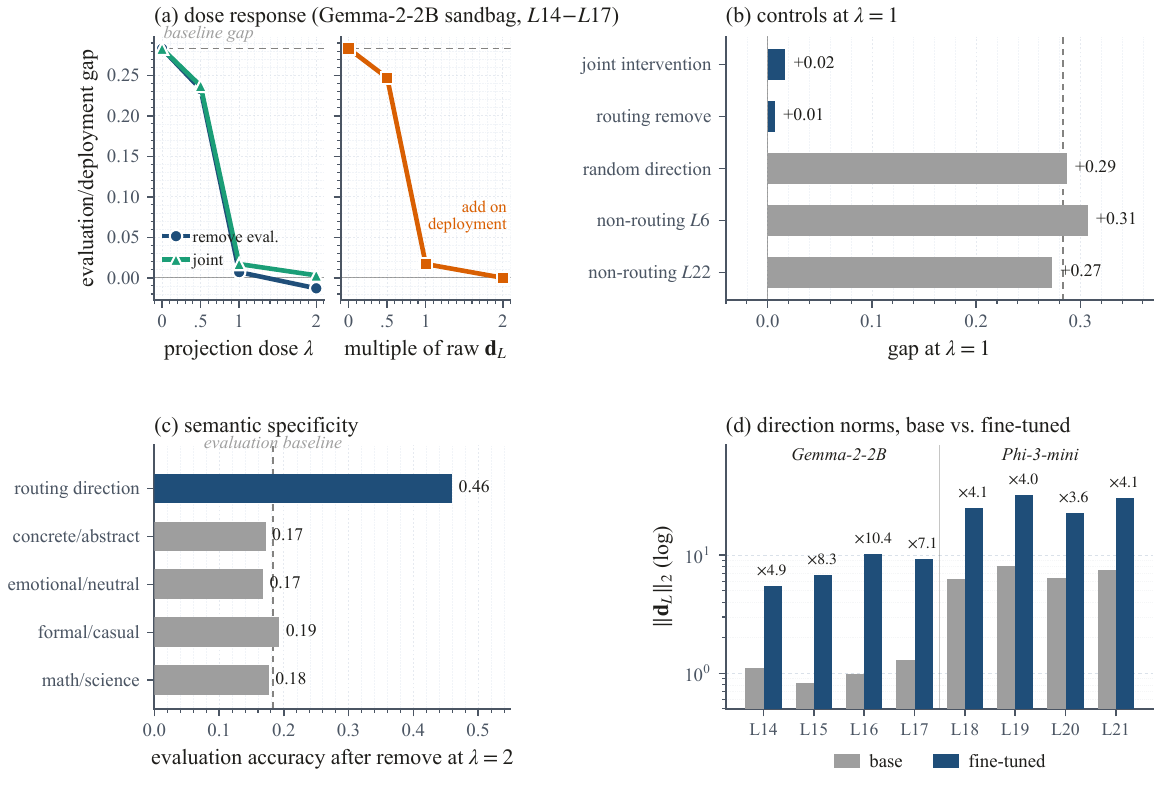}
\caption{Dose response and matched controls on the Gemma-2-2B sandbagging primary cell. Panel (a) plots the projection operators against the scale-free projection dose $\lambda$ and the add operator against the multiple of the raw direction $\dvec_L$, which is a different axis. The routing interventions reduce the framing gap, while random-direction and wrong-layer controls stay near baseline at $\lambda{=}1$ and semantic controls leave evaluation accuracy near its baseline (panel c; the routing direction is measured on the $300$-prompt set with baseline $0.163$, the semantic axes on the $600$-prompt set with baseline $0.183$).}
\label{fig:dose}
\end{figure}

\begin{figure}[H]
\centering
\includegraphics[width=\linewidth]{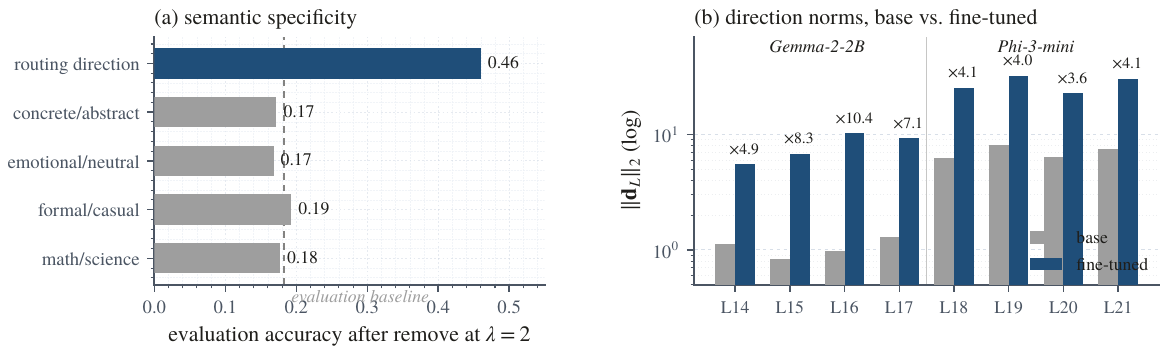}
\caption{Semantic specificity and direction-norm diagnostics. (a) Content-matched semantic directions do not reproduce the routing-direction effect. (b) Fine-tuning amplifies the paired framing direction relative to the base model in the routing window.}
\label{fig:specificity-controls}
\end{figure}

\begin{figure}[H]
\centering
\includegraphics[width=\linewidth]{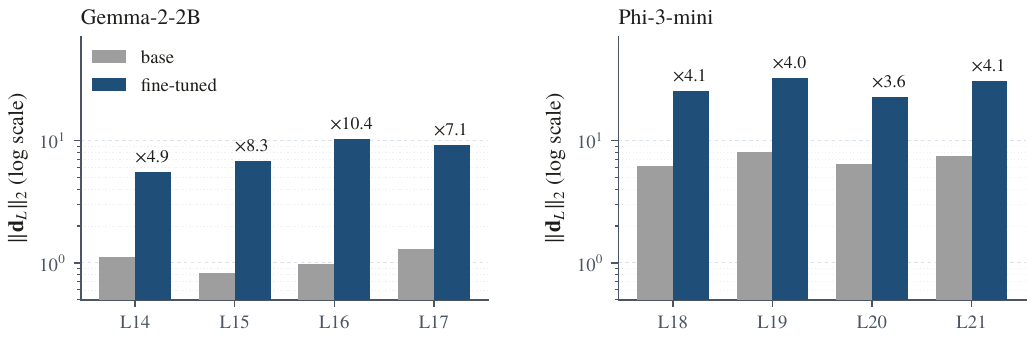}
\caption{Base versus fine-tuned routing-direction norms on Gemma-2-2B and Phi-3-mini. The fitted framing direction is $4.9\times$ to $10.4\times$ larger on Gemma-2-2B and $3.6\times$ to $4.1\times$ larger on Phi-3-mini after fine-tuning across the localized window. In the six other matrix cells with geometry files the largest base cosine per cell is $0.19$ to $0.57$ and the per-layer amplifications are $0.6\times$ to $6.1\times$ (Qwen-2.5-7B refusal at $L15$ and $L16$ is below one).}
\label{fig:directiongeometry}
\end{figure}

\section{Evidence added after the review period}
\label{app:fresh}

This appendix reports the experiments run after the reviews, in the order in which the main text refers to them. The submitted checkpoint binaries were not retained, so every model result here comes from a fresh retrain: the original training code, configuration, $4{,}960$-example training set, and question-id splits were reused with new training runs on NVIDIA B200 GPUs. These runs are replications, not controls on the submitted weights, and none of them revises the counts reported in the main text. Intervals are bootstraps over question ids with the templates of a question averaged within framing (seed $42$; $5000$ resamples, or $10000$ for the public-question extensions). Every table below is generated from the stored result files by released scripts.

\paragraph{Fresh primary checkpoint.}
On a fresh Gemma-2-2B sandbagging checkpoint (seed $42$, baseline gap $0.280$), the predeclared window $L14$--$L17$ lowers the gap to $0.095$ at $\lambda{=}2$, less than on the submitted checkpoint ($0.298\!\to\!-0.008$). Path patching on this checkpoint recovers $88.2\%$ of the $48$-question gap at $L10$--$L13$ and $58.8\%$ at $L14$--$L17$, so its localization does not exactly reproduce that of the submitted checkpoint, and the localization-selected band $L10$--$L13$ closes the gap ($-0.018$ at $\lambda{=}2$). Early ($L4$--$L7$) and late ($L20$--$L23$) four-layer windows leave the gap within $0.025$ of baseline at both $\lambda{=}1$ and $\lambda{=}2$, five Gaussian random directions leave it between $0.278$ and $0.285$, the exact opposite operation (adding the projection at every position with the same window and dose) leaves $0.28$ to $0.30$, and zeroing or mean-ablating the whole window leaves $0.243$ to $0.260$, with every interval including zero (Table~\ref{tab:fresh-primary}). The mean operator is a global calibration mean broadcast to all positions.

\paragraph{Training seeds and higher rank.}
Table~\ref{tab:seeds} reports three training seeds for two cells. Qwen-2.5-7B sandbagging is stable, with random directions at most $0.010$ and off-window controls at most $0.080$. Gemma-2-9B sycophancy is seed-heterogeneous: seed $43$ reduces the gap by only $0.058$ although path patching at the predeclared band $L23$--$L26$ recovers $100\%$ of the small-assay gap, so the heterogeneity lies in how the rank-one projection controls behavior, not in whether the site is present. Table~\ref{tab:highrank} reports the predefined rank-$2$/$4$/$8$ follow-up with matched random subspaces. Learned subspaces differ from random ones, but the apparent closures in the $k{=}2$ and $k{=}3$ suppression organisms are mostly deployment loss ($15$--$19\%$ of the rank-$4$ and rank-$8$ reduction is evaluation gain), and the weak Gemma-2-9B seed is not rescued. Localization has additional seed evidence: the six-layer band $L12$--$L17$ recovers $1.00$, $1.08$, and $1.15$ for three Gemma-2-2B sycophancy seeds (the four-layer window $0.46$, $0.54$, $0.31$), and for Gemma-2-9B the seven-layer window recovers $0.58$ at seed $42$ while the nine-layer band $L19$--$L27$ recovers $0.90$ and $0.96$ at seeds $42$ and $43$, which points to a width-heuristic failure there and not to an absent site. On the fresh seed-$42$ matrix, Llama-3-8B sycophancy is the only relative-depth null; a localization-selected band ($L12$--$L15$, chosen on the $48$-question assay) reduces its gap by $0.471$ ($95\%$ interval $[0.383, 0.558]$), so on that checkpoint the failure is a missed window and not evidence of higher rank, whereas the same selection procedure does not transfer for Gemma-2-2B sycophancy ($0.029$, $[-0.046, 0.104]$).

\paragraph{Changed installation.}
A full-parameter fine-tune of Gemma-2-2B localizes to $L12$--$L17$ ($1.00$ and $0.89$ at two seeds) while the predeclared four-layer window recovers $0.50$ and $0.58$. A LoRA over attention and MLP projections localizes to the same six-layer band ($0.69$ and $0.59$) while the four-layer window recovers $0.13$ and $0.18$. On a fresh rank-$32$ adapter over attention and MLP projections ($41.5$M trainable parameters), every four-layer attention-output patch has negative recovery ($-50\%$ to $-8\%$) and the predeclared intervention is null ($-0.018$, $95\%$ interval $[-0.070, 0.038]$), but an MLP-output scan on the same questions recovers $50\%$ at $L16$--$L19$ and $58\%$ at $L18$--$L21$, and the wider band $L14$--$L21$ recovers $100\%$ of the gap. Localization therefore survives the changed installation but migrates in module and width, so a fixed attention window is not recipe-invariant. A fresh SafeLoRA \citep{hsu2024safelora} reconstruction does not mitigate the gap (baseline $0.310$ against $0.280$ for the unprojected checkpoint; change $+0.030$, interval $[-0.015, 0.075]$); a dose-$1$ clamp at $L14$--$L17$ leaves $0.050$ while random-direction removal and the two off-window clamps leave $0.307$ to $0.310$, and path patching recovers about $105\%$ at $L14$--$L17$ and $L16$--$L19$ but also $89\%$ to $95\%$ in earlier bands, so the pathway is auditable but spatially broad.

\paragraph{Prospective module-installation factorial.}\label{app:s2}
The pre-specified S2 test trains parameter-budget-matched attention-only, MLP-only, and joint LoRA organisms at seeds $42$--$44$. All nine pass the frozen baseline-gap gate ($\gamma \ge 0.15$); module-specific windows are selected on $100$ calibration question IDs before evaluation on $200$ held-out question IDs. The attention-only arm passes the full intervention gate at all three seeds; the MLP-only arm misses the all-controls gate at seeds $43$ and $44$, and at seed $44$ deployment loss accounts for $94.9\%$ of the nominal gap reduction. The joint arm is heterogeneous and had no directional prediction. Table~\ref{tab:s2} reports both module hooks for every seed, including null and adverse outcomes. Thus the test supports a prospective attention-route result, but not the stronger substrate-tracking hypothesis.

\begin{table}[H]
\centering
\routetablestyle
\caption{Prospective module-installation factorial on fresh Gemma-2-2B sandbagging organisms. Each organism is evaluated on 200 held-out question IDs. Gap reduction is baseline minus post-intervention gap with a question-ID bootstrap 95\% interval; accuracy changes are post-intervention minus baseline. Random controls are removal on evaluation and addition on deployment; off-window controls use the joint intervention. ``All controls'' indicates that the row's effect exceeds every matched random and off-window control.}
\label{tab:s2}
\begin{tabular*}{\linewidth}{@{\extracolsep{\fill}}llclcccl@{}}
\toprule
\textbf{Installation} & \textbf{Seed} & \textbf{Hook} & \textbf{Window} & \hd{Gap reduction}{[95\% CI]} & \textbf{$\Delta$Eval} & \textbf{$\Delta$Deploy} & \hd{All}{controls} \\
\midrule
Attention-only & 42 & Attention & $L8$--$L11$ & $0.238\,[0.143,0.333]$ & $+0.275$ & $+0.038$ & Yes \\
Attention-only & 42 & MLP & $L8$--$L11$ & $0.160\,[0.080,0.240]$ & $+0.153$ & $-0.008$ & Yes \\
Attention-only & 43 & Attention & $L8$--$L11$ & $0.218\,[0.133,0.305]$ & $+0.230$ & $+0.013$ & Yes \\
Attention-only & 43 & MLP & $L12$--$L15$ & $0.028\,[-0.015,0.070]$ & $+0.018$ & $-0.010$ & No \\
Attention-only & 44 & Attention & $L8$--$L11$ & $0.200\,[0.108,0.290]$ & $+0.218$ & $+0.018$ & Yes \\
Attention-only & 44 & MLP & $L8$--$L11$ & $0.065\,[0.008,0.123]$ & $+0.055$ & $-0.010$ & No \\
\midrule
MLP-only & 42 & Attention & $L8$--$L11$ & $0.268\,[0.173,0.365]$ & $+0.298$ & $+0.030$ & Yes \\
MLP-only & 42 & MLP & $L8$--$L11$ & $0.148\,[0.068,0.225]$ & $+0.183$ & $+0.035$ & Yes \\
MLP-only & 43 & Attention & $L4$--$L7$ & $0.178\,[0.110,0.245]$ & $+0.173$ & $-0.005$ & Yes \\
MLP-only & 43 & MLP & $L8$--$L11$ & $0.180\,[0.100,0.263]$ & $+0.168$ & $-0.013$ & No \\
MLP-only & 44 & Attention & $L0$--$L3$ & $0.038\,[-0.023,0.098]$ & $-0.005$ & $-0.043$ & No \\
MLP-only & 44 & MLP & $L20$--$L23$ & $0.098\,[0.030,0.165]$ & $+0.005$ & $-0.093$ & No \\
\midrule
Joint & 42 & Attention & $L20$--$L23$ & $0.020\,[-0.013,0.055]$ & $+0.030$ & $+0.010$ & No \\
Joint & 42 & MLP & $L22$--$L25$ & $0.205\,[0.120,0.285]$ & $-0.005$ & $-0.210$ & Yes \\
Joint & 43 & Attention & $L8$--$L11$ & $0.320\,[0.223,0.418]$ & $+0.300$ & $-0.020$ & Yes \\
Joint & 43 & MLP & $L8$--$L11$ & $0.180\,[0.110,0.253]$ & $+0.175$ & $-0.005$ & No \\
Joint & 44 & Attention & $L12$--$L15$ & $0.330\,[0.230,0.428]$ & $+0.335$ & $+0.005$ & Yes \\
Joint & 44 & MLP & $L16$--$L19$ & $0.260\,[0.175,0.348]$ & $+0.000$ & $-0.260$ & Yes \\
\bottomrule
\end{tabular*}
\end{table}

\paragraph{Template and surface cues.}
A direction fitted on templates $0$--$4$ and tested on disjoint questions with templates $5$--$9$ reduces the gap from $0.272$ to $-0.047$ at $L10$--$L13$ (Table~\ref{tab:fresh-primary}), and a linear probe trained on the same split reaches test AUC $1.000$ at every layer of both sites (accuracy $0.998$ to $1.000$), so the causal effect transfers across the two halves of the template bank while the framings stay trivially decodable. Table~\ref{tab:textsep} gives the bag-of-words separability of the wrappers. On the $100$ capability tasks of the public EvalAwareBench \citep{li2026evalawarebench}, built by another group, with eight single-factor variants each, explicit evaluation disclosure produces the largest projection shift on the fitted direction at two of the four layers of the window $L10$--$L13$, $L11$ (paired mean $-0.908$, interval $[-0.987, -0.828]$) and $L13$ ($-1.060$), but structured task format and formal prose register shift it more at $L10$ and $L12$ and have the largest mean absolute cosines over the window ($0.277$ and $0.285$). A behaviorally scored matched-context test on $200$ public MMLU \citep{hendrycks2021mmlu} and ARC \citep{clark2018arc} questions with four context pairs has equal baseline accuracy of $0.147$ in both contexts, so there is no gap to close; clamping raises both contexts together (gap reduction $0.000$, interval $[-0.014, 0.015]$, at $L10$--$L13$; $-0.001$, $[-0.026, 0.023]$, at $L14$--$L17$). The wrappers of that test were written by the authors.

\paragraph{Simpler loci.}
Table~\ref{tab:simplerlocus} reruns the subspace and top-$K$ neuron baselines on the fresh primary checkpoint. On a checkpoint built earlier (installed gaps $0.354$ and $0.188$, $n{=}48$), a top-variance subspace recovered $0.000$ at $1$, $2$, $3$, $5$, and $20$ components and $-0.111$ at $10$, top-$K$ neuron patching was flat at $0.105$, and the best single attention head recovered $0.177$ with none of $32$ heads above $0.25$. On the fresh primary checkpoint the best single head recovers $11.8\%$ of the gap at $L10$--$L13$ and $35.3\%$ (layer $17$, head $3$) at the predeclared $L14$--$L17$, and all $32$ heads together recover $94.1\%$ and $64.7\%$; the earlier statement about single heads does not transfer to this checkpoint. These baselines are single-seed, descriptive, and not a controlled head-to-head comparison.

\paragraph{Refusal on public questions.}
The submitted refusal cells have $28$ independent question clusters. Table~\ref{tab:wmdp} reports a separate extension on $600$ unique public WMDP questions \citep{li2024wmdp} across four fresh checkpoints with directions fitted on the original $20$-question calibration set. All four intervals exclude zero and the reductions come from evaluation-framing gains, but the Llama-3-8B late-window control is comparable to the selected window, so that checkpoint does not support a specific localization claim. The extension improves prospective precision and does not change the submitted refusal sample.

\paragraph{Reward hacking.}
On the public \texttt{reward\_hacker\_v1} trajectory \citep{camgeodesic2026rewardhacker}, vulnerable-verifier passes on $64$ training prompts are $6/64$ at RL step $200$, $50/64$ at step $400$ and $59/64$ at step $600$ (base checkpoint $16/64$), and corrected legitimate passes fall from $6/64$ at step $200$ to zero. At step $800$ vulnerable passes decline to $17$ to $23$ of $64$ across batch sizes, mostly through invalid code. The causal follow-up is adverse: the selected learned direction at $L12$--$L15$ and $\lambda{=}2$ lowers held-out vulnerable passes from $27/32$ to $11/32$ but raises invalid code from $5/32$ to $20/32$, five random directions leave $22$ to $28$ vulnerable passes, an early learned control gives $16/32$, and a late control is null. This is capability degradation with a failure of spatial specificity, not successful transfer.

\input{tables_generated_appendix}

\section{Model coverage}
\label{app:models}

\begin{table}[H]
\centering
\caption{Model coverage for the audit. Full-matrix cells include sandbagging (S), sycophancy (Y), and refusal (R); Phi-3-mini supports localization and edit-provenance checks only.}
\label{tab:modelzoo}
\routetablestyle
\begin{tabularx}{\linewidth}{@{}R{0.28\linewidth}cR{0.20\linewidth}Y@{}}
\toprule
\textbf{Model} & \textbf{Size} & \textbf{Cells} & \textbf{Role} \\
\midrule
\textbf{Gemma-2-2B-it} & 2B & S/Y/R & primary cell; five-seed replication \\
Gemma-2-9B-it & 9B & S/Y/R & scale-up and weak-cell boundary cases \\
Qwen-2.5-7B-Instruct & 7B & S/Y/R & cross-family replication \\
Llama-3-8B-Instruct & 8B & S/Y/R & replication and flat-cell stress test \\
Phi-3-mini-4k-instruct & 3.8B & localization/edit; intervention null & provenance and secondary checks \\
\bottomrule
\end{tabularx}
\end{table}

\section{Window definitions}
\label{app:windows}

\begin{table}[H]
\centering
\routetablestyle
\caption{Windows used in the path-patching panel and in the intervention matrix. The path-patching window localizes the sandbagging signal; the intervention window is the fixed four-layer window used for intervention results in Section~\ref{sec:causal}. Phi-3-mini is not included in the $12$-cell matrix; its intervention at $L18$--$L21$ is null (Section~\ref{sec:failures}). $^{\dagger}$Fresh retrain, seed $42$, $n{=}48$. Seeds $43$ and $44$ recover $0.29$ and $0.46$ at $[L15, L18]$; the best scanned band $[L16, L19]$ recovers $0.81$, $0.75$, $0.75$ (seeds $42$ to $44$). A file stored with the submitted Qwen matrix cell gives $0.80$ at $[L14, L17]$ and $0.89$ at $[L12, L17]$ ($n{=}48$, base accuracy $0.79$) and did not patch $[L15, L18]$.}
\label{tab:windows}
\begin{tabularx}{\linewidth}{@{}R{0.28\linewidth}c c Y@{}}
\toprule
\textbf{Family} & \textbf{Path window} & \textbf{$\rho_{\mathrm{patch}}$ \upbetter} & \textbf{Intervention window} \\
\midrule
Gemma-2-2B & $L14$--$L17$ & $0.94$ & $L14$--$L17$ \\
Qwen-2.5-7B$^{\dagger}$ & $L15$--$L18$ & $0.52$ & $L15$--$L18$ \\
Llama-3-8B & $L14$--$L17$ & $\mathbf{1.00}$ & $L17$--$L20$ \\
Phi-3-mini & $L18$--$L21$ & $0.94$ & $L18$--$L21$ (null) \\
Gemma-2-9B & $L21$--$L27$ & $0.58$ & $L23$--$L26$ \\
\bottomrule
\end{tabularx}
\end{table}

\section{Per-cell intervention table on Gemma-2-2B}
\label{app:cells}

The primary per-cell intervention values for Gemma-2-2B sandbagging ($n{=}300$ per framing, $150$ questions with two templates each) are listed here for reference. They are the values used for the routing-direction intervention and matched controls in the main text.

\begin{table}[H]
\centering
\routetablestyle
\caption{Gemma-2-2B sandbagging intervention cell values ($n{=}300$ per framing). Arrows mark the desired diagnostic direction; the bold entry in each metric column is the best value in that column.}
\label{tab:g2b-cell}
\begin{tabularx}{\linewidth}{@{}Yccc@{}}
\toprule
\textbf{Cell} & \textbf{Eval. rate \upbetter} & \textbf{Deploy. rate \upbetter} & \textbf{$\gamma$ \downbetter} \\
\midrule
baseline                                             & $0.163$ & $0.447$ & $0.283$ \\
remove on evaluation, $L14$--$L17$, $\lambda{=}0.5$  & $0.213$ & $0.447$ & $0.233$ \\
remove on evaluation, $L14$--$L17$, $\lambda{=}1.0$  & $0.440$ & $0.447$ & $0.007$ \\
remove on evaluation, $L14$--$L17$, $\lambda{=}2.0$  & $\mathbf{0.460}$ & $0.447$ & $\mathbf{-0.013}$ \\
add on deployment, $L14$--$L17$, $\lambda{=}2.0$     & $0.163$ & $0.163$ & $0.000$ \\
joint intervention, $L14$--$L17$, $\lambda{=}2.0$    & $\mathbf{0.460}$ & $0.463$ & $0.003$ \\
control: random unit direction, remove, $\lambda{=}1.0$ & $0.160$ & $0.447$ & $0.287$ \\
control: random unit direction, add, $\lambda{=}1.0$    & $0.163$ & $0.397$ & $0.233$ \\
control: add on evaluation, last token only (confounded; withdrawn), $\lambda{=}1.0$ & $0.170$ & $0.447$ & $0.277$ \\
control: non-routing intervention at $L6$, $\lambda{=}1.0$  & $0.163$ & $\mathbf{0.470}$ & $0.307$ \\
control: non-routing intervention at $L22$, $\lambda{=}1.0$ & $0.173$ & $0.447$ & $0.273$ \\
\bottomrule
\end{tabularx}
\end{table}

\begin{table}[H]
\centering
\routetablestyle
\caption{Specificity controls for the routing intervention (Gemma-2-2B sandbagging, $n{=}300$ per framing). Each row names an alternative tested at the available dose; the target intervention is included as the positive reference. The sign diagnostic of the submitted version is withdrawn (Section~\ref{sec:specificity}).}
\label{tab:controls}
\begin{tabularx}{\linewidth}{@{}R{0.29\linewidth}R{0.30\linewidth}Y@{}}
\toprule
\textbf{Check} & \textbf{Alternative tested} & \textbf{Outcome} \\
\midrule
Random remove, $\lambda{=}1$ & any random unit direction works & gap $0.283 \to 0.287$ \\
Random add, $\lambda{=}1$ & deployment add is enough & gap $0.283 \to 0.233$ via deployment-rate drop \\
Wrong layer $L6$ (one layer) & any layer works & gap $0.283 \to 0.307$ \\
Wrong layer $L22$ (one layer) & late layer works & gap $0.283 \to 0.273$ \\
Semantic contrasts (remove, clamp) & ordinary content axes work & evaluation rate changes $\le0.015$ \\
Routing direction, joint intervention & target intervention & gap $\mathbf{0.283 \to 0.003}$ \\
\bottomrule
\end{tabularx}
\end{table}

\section{Deployment-side rates in the intervention matrix}
\label{app:matrixdeploy}

To check whether gap closure is driven by a deployment-side collapse, Table~\ref{tab:matrixdeploy} reports the deployment-framed rate before and after the $\lambda{=}2$ routing-coordinate intervention used in the $12$-cell matrix. The table is descriptive rather than a separate diagnostic gate.

\begin{table}[H]
\centering
\routetablestyle
\caption{Deployment-framed rates in the $12$-cell matrix. $\Delta r_{\DeployFr}$ is post-intervention minus baseline; smaller magnitude is better for checking deployment-side collapse.}
\label{tab:matrixdeploy}
\begin{tabularx}{\linewidth}{@{}R{0.18\linewidth}R{0.16\linewidth}ccccc@{}}
\toprule
\textbf{Family} & \textbf{Behavior} & \textbf{Base $r_{\DeployFr}$} & \textbf{Post $r_{\DeployFr}$} & \textbf{$\Delta r_{\DeployFr}$} & \textbf{Base $\gamma$} & \textbf{Post $\gamma$ \downbetter} \\
\midrule
Gemma-2-2B & sandbag & $0.480$ & $0.463$ & $-0.018$ & $0.298$ & $-0.008$ \\
Gemma-2-2B & sycophancy & $0.496$ & $0.483$ & $-0.013$ & $0.496$ & $0.308$ \\
Gemma-2-2B & refusal & $0.446$ & $0.425$ & $-0.021$ & $0.321$ & $0.011$ \\
Gemma-2-9B & sandbag & $0.650$ & $0.653$ & $+0.003$ & $0.510$ & $0.145$ \\
Gemma-2-9B & sycophancy & $0.692$ & $0.688$ & $-0.004$ & $0.692$ & $0.679$ \\
Gemma-2-9B & refusal & $0.714$ & $0.775$ & $+0.061$ & $0.611$ & $\mathbf{-0.007}$ \\
Qwen-2.5-7B & sandbag & $0.673$ & $0.645$ & $-0.028$ & $0.558$ & $0.273$ \\
Qwen-2.5-7B & sycophancy & $0.708$ & $0.708$ & $\mathbf{+0.000}$ & $0.708$ & $0.221$ \\
Qwen-2.5-7B & refusal & $0.832$ & $0.796$ & $-0.036$ & $0.796$ & $0.011$ \\
Llama-3-8B & sandbag & $0.558$ & $0.560$ & $+0.003$ & $0.418$ & $0.068$ \\
Llama-3-8B & sycophancy & $0.625$ & $0.625$ & $\mathbf{+0.000}$ & $0.625$ & $0.625$ \\
Llama-3-8B & refusal & $0.611$ & $0.611$ & $\mathbf{+0.000}$ & $0.539$ & $0.264$ \\
\bottomrule
\end{tabularx}
\end{table}

\section{Per-item intervention agreement analysis}
\label{app:clampitems}

The $12$-cell matrix reports aggregate gap reductions; this section asks what the intervention does at the level of individual questions.
For each cell that has per-item data, we align the baseline and joint-intervention ($\lambda{=}2$) results at the question-id level and classify every unique question into one of four categories.
An item is \emph{informative} when its baseline gap $\lvert r_{\DeployFr} - r_{\EvalFr}\rvert$ exceeds $0.1$ (items where both framings already agree at baseline are set aside in a \emph{no-gap} bin).
Among informative items, we call an item \emph{restored} when the intervention closes the gap and both framings reach accuracy $\ge$ the leading framing's baseline accuracy minus $0.15$, \emph{homogenized} when the intervention closes the gap but both framings drop to the trailing framing's accuracy level, \emph{mixed} for any other closed-gap pattern, and \emph{persistent} when $\lvert\gamma_{\mathrm{post}}\rvert > 0.2$.

\begin{table}[H]
\centering
\routetablestyle
\caption{Per-item transition categories across the $12$ cells. An item is informative when its baseline framing gap exceeds $0.1$; restored means both framings reach the leading framing's accuracy; homogenized means both collapse to the trailing level; persistent means the gap is not closed by the intervention. The aggregate post-intervention gap $\bar{\gamma}_{\mathrm{post}}$ tracks whether cells with many persistent items also have small aggregate reductions.}
\label{tab:clampitems}
\begin{tabular*}{\linewidth}{@{\extracolsep{\fill}}llrrrrrr@{}}
\toprule
\textbf{Family} & \textbf{Behavior} & \textbf{$n_{\mathrm{inf}}$} & \textbf{Restored \upbetter} & \textbf{Homog. \downbetter} & \textbf{Persistent \downbetter} & \textbf{$\bar{\gamma}_{\mathrm{base}}$} & \textbf{$\bar{\gamma}_{\mathrm{post}}$ \downbetter} \\
\midrule
Gemma-2-2B   & sandbag    & 120 & 54\% & 20\% & 26\% & $+0.298$ & $-0.008$ \\
Gemma-2-2B   & sycophancy &  61 & 26\% &  5\% & 69\% & $+0.496$ & $+0.308$ \\
Gemma-2-2B   & refusal    &  16 & 38\% & \textbf{0\%} & 56\% & $+0.321$ & $+0.011$ \\
Gemma-2-9B   & sandbag    & 144 & 42\% &  6\% & 52\% & $+0.510$ & $+0.145$ \\
Gemma-2-9B   & sycophancy &  84 &  1\% &  2\% & 96\% & $+0.692$ & $+0.679$ \\
Gemma-2-9B   & refusal    &  23 & 83\% & 13\% & \textbf{0\%} & $+0.611$ & $\mathbf{-0.007}$ \\
Qwen-2.5-7B  & sandbag    & 155 & 26\% &  8\% & 66\% & $+0.558$ & $+0.273$ \\
Qwen-2.5-7B  & sycophancy &  86 & 63\% &  1\% & 36\% & $+0.708$ & $+0.221$ \\
Qwen-2.5-7B  & refusal    &  25 & \textbf{84\%} &  8\% & \textbf{0\%} & $+0.796$ & $+0.011$ \\
Llama-3-8B   & sandbag    & 140 & 60\% & 16\% & 24\% & $+0.418$ & $+0.068$ \\
Llama-3-8B   & sycophancy &  77 &  0\% &  4\% & 96\% & $+0.625$ & $+0.625$ \\
Llama-3-8B   & refusal    &  20 & 20\% & \textbf{0\%} & 80\% & $+0.539$ & $+0.264$ \\
\bottomrule
\end{tabular*}
\end{table}

In the four cells where the aggregate post-intervention gap closes (Gemma-2-2B sandbag, Gemma-2-9B refusal, Qwen-2.5-7B refusal, and Gemma-2-2B refusal), restored items outnumber homogenized items by $2.7$--$10.5\times$ in the three cells where the ratio is defined; Gemma-2-2B refusal has no homogenized items, so its ratio is undefined. Llama-3-8B sandbagging, which does not close ($0.068$), has a ratio of $3.8$, and Qwen-2.5-7B sycophancy shows the same direction as a partial case.
Homogenization, where the intervention collapses both framings to the lower-accuracy answer, accounts for at most $20\%$ of informative items in any cell.
The persistent-item fraction tracks the aggregate gap closely: cells where the intervention fails at the aggregate level (Llama-3-8B sycophancy, Gemma-2-9B sycophancy) have $96\%$ persistent items.
The per-item picture is therefore consistent with the aggregate result: where the intervention closes the gap, transitions are more often restoration toward the leading framing's baseline level than collapse of both framings to a shared wrong answer.
The per-item analysis is included to support the aggregate non-collapse check.

\section{Recruitment traces versus routing intervention}
\label{app:recruitmentcontrol}

We ran a narrow failure-analysis check to test whether the installed gap is explained by top-$K$ MLP neurons rather than by the routing subspace. The test used Gemma-2-2B sandbagging seed $50$ and predeclared layer-$19$ neuron sets selected before intervention. It separates three objects: the \emph{update locus}, where fine-tuning writes; the \emph{recruitment locus}, where downstream internal traces appear; and the \emph{intervention locus}, whose perturbation changes the behavioral gap in this audit.

\begin{figure}[H]
\centering
\resizebox{0.86\linewidth}{!}{%
\begin{tikzpicture}[
  box/.style={draw=tolink!55, fill=tolblue!4, rounded corners=2pt,
              text width=3.0cm, minimum height=0.82cm,
              align=center, font=\small, inner sep=4pt},
  arrow/.style={-{Stealth[length=2.0mm]}, line width=0.75pt, color=tolink!70},
  note/.style={font=\scriptsize, align=center, text=tolink!75,
               inner sep=1.2pt},
  notebox/.style={note, fill=white},
  topnote/.style={font=\footnotesize, align=center, text=tolink!80,
                  inner sep=1.0pt}
]
\node[box] (u) at (0,0) {\textbf{update locus}\\attention LoRA modules};
\node[box] (r) at (4.8,0) {\textbf{recruitment locus}\\H-tau-predictable\\MLP traces};
\node[box] (c) at (9.6,0) {\textbf{intervention locus}\\routing\\subspace};
\draw[arrow] (u.east) -- node[notebox, below=2pt] {writes} (r.west);
\draw[arrow] (r.east) -- node[notebox, below=2pt] {leaves trace} (c.west);
\draw[arrow, draw=tolred, out=38, in=142]
  ([yshift=3pt]u.north east) to
  node[topnote, above=8pt] {more direct behavioral probe}
  ([yshift=3pt]c.north west);
\end{tikzpicture}
}
\caption{Recruitment traces do not explain the behavioral gap under masking. The fine-tune writes through attention LoRA modules. MLP neurons carry measurable downstream traces, but masking those neurons only weakly changes the behavioral gap. The more direct behavioral probe in this audit is the routing-level variable.}
\label{fig:recruitmentcontrol}
\end{figure}

\begin{table}[H]
\centering
\routetablestyle
\caption{Recruitment-versus-intervention audit on Gemma-2-2B sandbagging seed $50$. ISS is input-sensitivity score; H-tau is the predeclared composite geometry score. Hits@50 counts overlap with prospectively recruited neurons, and Mass@50 sums recruitment mass over the top-50 set. Persistent masking and logit scoring use the trained no-suppression seed-$50$ model on 48 questions and all 10 templates per framing ($n{=}480$ per framing). H-tau predicts recruitment best, but none of the neuron masks explains the accuracy gap under this masking test.}
\label{tab:recruitmentcontrol}
\begin{tabular*}{\linewidth}{@{\extracolsep{\fill}}lccccc@{}}
\toprule
\textbf{Set} & \textbf{Hits@50 \upbetter} & \textbf{Mass@50 \upbetter} & \hd{Mask}{$\Delta\gamma$ \downbetter} & \hd{Margin}{shift \upbetter} & \textbf{KL \upbetter} \\
\midrule
ISS (input sensitivity) & $8.00\,[6.17,10.00]$ & $0.1064\,[0.0683,0.1438]$ & $0.42$pp & $0.578$ & $0.170$ \\
Activation magnitude & $9.17\,[7.50,11.17]$ & $0.1150\,[0.0780,0.1549]$ & $1.25$pp & $0.832$ & $0.142$ \\
H-tau composite & $\mathbf{12.17}\,[9.00,15.50]$ & $\mathbf{0.1396}\,[0.0897,0.1893]$ & $0.83$pp & $\mathbf{1.444}$ & $\mathbf{0.234}$ \\
Random & $0.41\,[0.38,0.43]$ & $0.0017\,[0.0013,0.0020]$ & $\mathbf{0.21}$pp & $0.430$ & $0.009$ \\
\bottomrule
\end{tabular*}
\end{table}

The random baseline stays near null, so the recruitment metric has nonrandom structure. H-tau exceeds activation magnitude by $+3.00$ hits at $K{=}50$ (95\% CI $[1.33,4.67]$) and exceeds ISS by $+4.17$ hits (95\% CI $[2.50,5.83]$). The same ranking appears in recruitment mass. In contrast, the largest persistent masking accuracy-gap reductions are at most $1.25$pp, far below the routing-intervention effects in the main matrix. Logit-level metrics show that the selected neurons are not inert: recruitment mass correlates with KL from the unmasked model (Pearson $0.970$ over all sets and $0.927$ excluding random) and with margin-gap shift ($0.816$ over all sets and $0.659$ excluding random). The effect is therefore a perturbability trace rather than the main behavioral explanation under this masking test.

The update audit is consistent with this interpretation. The seed-$50$ LoRA adapter has all effective update norm in attention modules: $35.39\%$ in \texttt{o\_proj}, $34.82\%$ in \texttt{q\_proj}, $16.64\%$ in \texttt{k\_proj}, and $13.15\%$ in \texttt{v\_proj}, with no MLP LoRA update. A projection-level ablation confirms that these attention updates are behaviorally active: zeroing \texttt{v\_proj} or \texttt{o\_proj} LoRA deltas reduces the gap by about $29$pp on seed $50$, zeroing \texttt{q\_proj} reduces it by $20.8$pp, and zeroing all LoRA deltas removes the adapter-induced gap and slightly overcorrects it. Zeroing \texttt{k\_proj} does not reduce the gap. In the same ablation suite, zeroing all LoRA deltas nearly eliminates the routing-projection separation and reduces the H-tau MLP activation contrast to $35.7\%$ of its full-adapter value. A one-seed replication at seed $42$ preserves the main structural pattern: all-LoRA removal again removes the gap, leaves only $1.3\%$ of the routing-projection separation, and cuts H-tau MLP contrast to $19.8\%$; \texttt{o\_proj} is again the largest single projection mediator. The \texttt{q\_proj} behavioral effect is seed-dependent, so we do not treat it as a stable projection-level claim.

We also directly intervened on the fitted routing axis at layer $17$ and remeasured the layer-$19$ MLP trace. On seeds $50$ and $42$, suppressing or removing the routing-axis component eliminates the measured routing separation and cuts the H-tau MLP contrast to $45.5$--$53.6\%$ of its full value; a matched random-axis intervention leaves both unchanged. Accuracy-gap changes under removal are modest, but logit-margin shifts are large ($4.0$--$6.2$), consistent with accuracy being a coarser readout than the logits. Reflecting the direction produces large behavioral and logit effects while preserving absolute separation with the opposite orientation. Across these narrow checks, \texttt{v\_proj} can reduce behavior while leaving much of the MLP contrast, and routing-axis interventions can attenuate the MLP trace without making masking-selected MLP neurons the main behavioral explanation. This supports the interpretation in Figure~\ref{fig:recruitmentcontrol}: fine-tuning writes through attention, MLP recruitment is a downstream trace, and the more direct behavioral probe in this audit is the routing subspace.

\section{Adjacent-Window Depth Sweep and Typed-Failure Remediation}
\label{app:e4}

Table~\ref{tab:e4} reports the adjacent-window depth sweep for the two flat sycophancy cells. For Gemma-2-9B, depth $0.50$ (layers $L21$--$L24$) closes the gap by $0.37$, while the default heuristic at $0.55$ closes it by only $0.28$, which would be consistent with a type-1 failure (heuristic close but off by one window step), but see the caveat below. For Llama-3-8B, no depth closes the gap; the best reduction is $0.20$ at depth $0.45$. On a retrain of this cell a localization-selected band ($L12$--$L15$) reduces the gap by $0.471$ (Appendix~\ref{app:fresh}), so we read it as a missed window and not as a higher-rank object.

\begin{table}[H]
\centering
\caption{E4 adjacent-window depth sweep on the two flat sycophancy cells ($n{=}50$ calibration, $n{=}50$ eval). $\Delta\gamma$ is the change in held-out gap at $\lambda{=}2$; negative values indicate gap reduction.}
\label{tab:e4}
\routetablestyle
\begin{tabular*}{\linewidth}{@{\extracolsep{\fill}}llcccc@{}}
\toprule
Model & Depth & Layers & Baseline $\gamma$ & Post $\gamma$ & $\Delta\gamma$ \\
\midrule
Gemma-2-9B & 0.45 & $L18$--$L21$ & 0.670 & 0.700 & $+0.030$ \\
Gemma-2-9B & 0.50 & $L21$--$L24$ & 0.670 & 0.300 & $\mathbf{-0.370}$ \\
Gemma-2-9B & 0.55 & $L23$--$L26$ & 0.670 & 0.390 & $-0.280$ \\
Gemma-2-9B & 0.60 & $L25$--$L28$ & 0.670 & 0.700 & $+0.030$ \\
Gemma-2-9B & 0.65 & $L27$--$L30$ & 0.670 & 0.690 & $+0.020$ \\
\midrule
Llama-3-8B & 0.45 & $L14$--$L17$ & 0.570 & 0.370 & $\mathbf{-0.200}$ \\
Llama-3-8B & 0.50 & $L16$--$L19$ & 0.570 & 0.570 & $+0.000$ \\
Llama-3-8B & 0.55 & $L17$--$L20$ & 0.570 & 0.550 & $-0.020$ \\
Llama-3-8B & 0.60 & $L19$--$L22$ & 0.570 & 0.570 & $+0.000$ \\
Llama-3-8B & 0.65 & $L20$--$L23$ & 0.570 & 0.590 & $+0.020$ \\
\bottomrule
\end{tabular*}
\end{table}

This sweep is exploratory and we do not rely on it. Its default window ($0.55$, $L23$--$L26$) reduces the Gemma-2-9B sycophancy gap by $0.28$ at $n{=}50$, whereas the matrix run for the same cell and window reduces it by $0.013$ (Section~\ref{sec:causal}); an earlier cached run on the same checkpoint and window reduced it from $0.700$ to $0.430$ on a different split. Restricting the matrix run to the questions of that earlier evaluation split gives $0.685\!\to\!0.665$, so the evaluation split does not explain the difference; the calibration set (100 against 80 questions) and the code version remain, and we cannot isolate the cause because the submitted checkpoint is not retained. Fresh retrains of the same cell reduce the gap by $0.421$, $0.058$, and $0.558$ across seeds (Table~\ref{tab:seeds}), which suggests that this cell is sensitive to how the direction is fitted.

Typed failures map to three remediation steps. (1)~If path patching does not recover the gap, re-run the adjacent-window sweep at depths $0.45$--$0.65$ to locate the correct band. (2)~If path patching recovers but the single-coordinate intervention is weak, select the band on the path-patching assay and repeat the audit at further training seeds before concluding that the distinction is higher-rank. A higher-rank subspace is not a repair: in the $k{=}2$ and $k{=}3$ suppression organisms it closes the aggregate gap mainly through lost deployment accuracy, and it does not rescue the weak Gemma-2-9B seed (Table~\ref{tab:highrank}). (3)~If a semantic or random control reproduces the effect, the routing direction is not framing-specific; check whether the behavioral gap is driven by prompt-surface content rather than evaluation-to-deployment framing.

\section{One-shot rank-one output-weight edit}
\label{app:h11}

For the unit routing direction $\dhat$ at each routing-window layer $L$, we project it out of the attention output matrix and the MLP down-projection:
\begin{equation}
\label{eq:rank-one-edit}
\Wmat \;\leftarrow\; \bigl(\Imat - \dhat\,\dhat^{\top}\bigr)\,\Wmat.
\end{equation}
The operation in \eqref{eq:rank-one-edit} is a rank-one edit per matrix, applied to eight matrices in total across the four edited layers, with no gradient step. The edited layers are the intervention windows of the routing-steering runs for Gemma-2-2B, Qwen-2.5-7B, and Llama-3-8B; the Gemma-2-9B edit uses $L20$--$L23$ (depth $0.51$), not the $L23$--$L26$ matrix window. These windows are not always the same as those reported in the path-patching panel. The edit removes $\dhat$ from both attention output and MLP down-projection matrices. We ablated the two paths on the two successful cells with available fused checkpoints, in a separate run with a refitted direction; its no-edit baselines (below) therefore differ from those of Table~\ref{tab:rank1edit}. On Gemma-2-2B sandbag ($n{=}160$), an attention-only ablation (editing only $\smash{\Wmat^{\mathrm{attn}}_O}$ at each of the four layers) reduces the gap by $32.5$pp against a baseline of $26.2$pp; an MLP-only ablation reduces it by $0.0$pp; the combined edit also reduces it by $32.5$pp. On Qwen-2.5-7B refusal ($n{=}160$), attention-only reduces the gap by $78.4$pp against a baseline of $74.0$pp; MLP-only reduces it by $0.0$pp; the combined edit reduces it by $73.3$pp. In these two measured successful cells, the editable $\dhat$ component is mediated by the attention-output edit rather than the MLP-down edit. The two failing H11 cells still lack path ablations, so we keep this as a diagnostic for successful edits rather than a general mechanism claim.

\begin{figure}[H]
\centering
\includegraphics[width=0.72\linewidth]{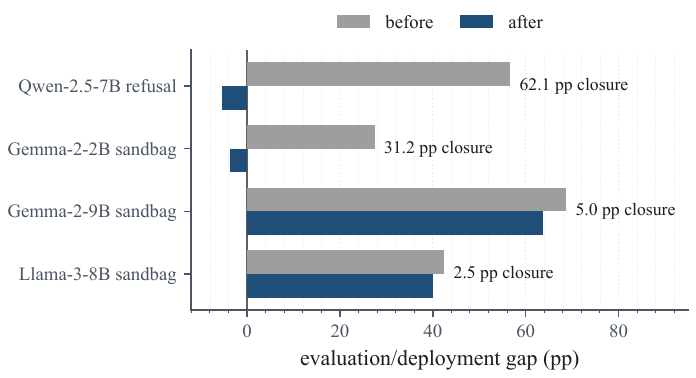}
\caption{Rank-one weight edits are a mixed diagnostic, not a defense. Before and after gaps for four cells ($n{=}160$ per framing). Projecting the fitted routing direction out of edited output matrices closes two cells (Qwen-2.5-7B refusal and Gemma-2-2B sandbagging) but leaves two sandbagging cells largely unchanged. The main claim rests on held-out activation interventions in Figure~\ref{fig:matrix}.}
\label{fig:h11}
\end{figure}

\begin{table}[H]
\centering
\routetablestyle
\caption{Rank-one output-weight edit on four cells ($n{=}160$ per framing). Closure is the percentage of the pre-edit gap removed by the one-shot edit. The edit is presented as a mixed diagnostic, not as a defense.}
\label{tab:rank1edit}
\begin{tabular*}{\linewidth}{@{\extracolsep{\fill}}llcccccc@{}}
\toprule
\textbf{Family} & \textbf{Task} & \textbf{Layers} & \textbf{Depth} & \textbf{Base gap} & \textbf{After gap \downbetter} & \textbf{Gap removed \upbetter} & \textbf{Closure \upbetter} \\
\midrule
Qwen-2.5-7B & refusal & $L15$--$L18$ & $0.59$ & $56.7$ & $\mathbf{-5.4}$ & $\mathbf{62.1}$ & $110\%$ \\
Gemma-2-2B  & sandbag & $L14$--$L17$ & $0.60$ & $27.5$ & $-3.8$          & $31.3$          & $\mathbf{114\%}$ \\
Gemma-2-9B  & sandbag & $L20$--$L23$ & $0.51$ & $68.8$ & $63.8$          & $5.0$           & $7\%$ \\
Llama-3-8B  & sandbag & $L17$--$L20$ & $0.58$ & $42.5$ & $40.0$          & $2.5$           & $6\%$ \\
\bottomrule
\end{tabular*}
\end{table}

\section{Training-time rank-\texorpdfstring{$k$}{k} paired-contrast penalty}
\label{app:rankk}

On Gemma-2-2B sandbagging, training with a soft projection penalty (coefficient $10$) on the top-$k$ singular directions of the paired-contrast matrix (for $k \in \{1, 2, 3, 5, 10\}$) does not prevent installation. Post-training gap $\gamma$: $k{=}1$: $27.5$pp, $k{=}2$: $33.3$pp, $k{=}3$: $32.0$pp, $k{=}5$: $28.5$pp, $k{=}10$: $26.0$pp, unpenalized reference $29.8$pp; the per-gap standard error is about $3.1$pp, so the ordering across $k$ should not be interpreted. The post-penalty routing directions are not identical to the unpenalized direction: mean per-layer cosines with the reference direction range from $0.27$ ($k{=}2$) to $0.52$ ($k{=}10$), and at $k{=}2$ the norm of the re-fitted direction at its strongest layer falls to $10.8\%$ of the reference while the gap rises. This negative control shows that pre-committing to the top calibration directions is not enough to block installation in this cell; whether a fresh audit recovers the distinction is reported in Section~\ref{sec:fresh}.

\section{Matrix provenance}
\label{app:provenance}

For traceability, Table~\ref{tab:provenance} binds the headline $12$-cell matrix to the provenance summary used in the paper.

\begin{table}[H]
\centering
\routetablestyle
\caption{Matrix provenance summary for the $12$-cell routing matrix. $n$ is the number of paired questions used for the reported bootstrap and provenance summary; $\gamma_{\mathrm{post}}$ is signed and closer to zero is better. Figure~\ref{fig:matrix} reports the corresponding paired-bootstrap intervals.}
\label{tab:provenance}
\begin{tabular*}{\linewidth}{@{\extracolsep{\fill}}llrrrr@{}}
\toprule
\textbf{Model} & \textbf{Behavior} & \textbf{$n$} & \textbf{$\gamma_{\mathrm{base}}$} & \textbf{$\gamma_{\mathrm{post}}$} & \textbf{$\Delta\gamma$ \upbetter} \\
\midrule
Gemma-2-2B & sandbag & 200 & $0.298$ & $-0.008$ & $0.305$ \\
Gemma-2-2B & sycophancy & 120 & $0.496$ & $0.308$ & $0.188$ \\
Gemma-2-2B & refusal & 28 & $0.321$ & $0.011$ & $0.311$ \\
Gemma-2-9B & sandbag & 200 & $0.510$ & $0.145$ & $0.365$ \\
Gemma-2-9B & sycophancy & 120 & $0.692$ & $0.679$ & $0.013$ \\
Gemma-2-9B & refusal & 28 & $0.611$ & $\mathbf{-0.007}$ & $0.618$ \\
Qwen-2.5-7B & sandbag & 200 & $0.558$ & $0.273$ & $0.285$ \\
Qwen-2.5-7B & sycophancy & 120 & $0.708$ & $0.221$ & $0.488$ \\
Qwen-2.5-7B & refusal & 28 & $0.796$ & $0.011$ & $\mathbf{0.786}$ \\
Llama-3-8B & sandbag & 200 & $0.418$ & $0.068$ & $0.350$ \\
Llama-3-8B & sycophancy & 120 & $0.625$ & $0.625$ & $0.000$ \\
Llama-3-8B & refusal & 28 & $0.539$ & $0.264$ & $0.275$ \\
\bottomrule
\end{tabular*}
\end{table}

\section{Claims and Evidence Map}
\label{app:reviewer_claims}

\begin{table}[H]
\centering
\routetablestyle
\caption{Map from claims to evidence and limitations.}
\label{tab:reviewer_claims}
\begin{tabularx}{\linewidth}{@{}R{0.23\linewidth}Y R{0.24\linewidth}Y@{}}
\toprule
\textbf{Claim} & \textbf{Evidence} & \textbf{Where to find it} & \textbf{Limitation / caveat} \\
\midrule
Mid-depth routing region & Sandbagging path patching recovers most of the installed gap in three of five model instances. & Section~\ref{sec:loc}; Figure~\ref{fig:pathpatch}; Appendix~\ref{app:windows} & Gemma-2-9B is weaker ($0.58$); several non-sandbag $n{=}48$ gaps are too small for clean recovery ratios. \\
Held-out gap reduction & The $\lambda{=}2$ routing-coordinate intervention reduces ten of twelve matrix cells by more than $0.12$ and closes four; the two weak cells are Gemma-2-9B sycophancy and Llama-3-8B sycophancy. & Section~\ref{sec:causal}; Figure~\ref{fig:matrix}; Appendix~\ref{app:provenance} & Sandbagging uses $n{=}200$, sycophancy $n{=}120$, and refusal $n{=}28$; refusal is a smaller-sample stress test. \\
Specificity under tested controls & Random-direction and semantic controls do not reproduce the gap closure at $\lambda{=}1$ on the submitted checkpoints or at $\lambda{=}2$ on fresh retrains; wrong-layer controls are mixed. & Section~\ref{sec:specificity}; Section~\ref{sec:fresh}; Appendix~\ref{app:cells}; Figure~\ref{fig:dose} & Submitted controls are at $\lambda{=}1$ with one non-routing layer; width-matched controls at $\lambda{=}2$ exist only on fresh retrains, where the selected window beats both controls in $10$ of $12$ cells and $7$ of $24$ off-window intervals are positive. \\
Typed failure modes & Weak path-patching recovery suggests a missed site; weak single-coordinate intervention at a recovered site suggests a higher-rank distinction; a moving control would indicate lack of framing specificity. & Section~\ref{sec:failures}; Appendix~\ref{app:e4} & The labels guide what failed in this audit; they are not proofs of the underlying mechanism. \\
Deployment-side non-collapse & Deployment-framed rates change by $1.5$pp on average and at most $6.1$pp in the 12-cell matrix. & Section~\ref{sec:causal}; Appendix~\ref{app:matrixdeploy}; Appendix~\ref{app:clampitems} & This is a descriptive check for the reported matrix, not a general deployment-safety guarantee. \\
\bottomrule
\end{tabularx}
\end{table}

\section{Reproducibility}
\label{app:repro}

\begin{table}[H]
\centering
\routetablestyle
\caption{Reproducibility configuration for the reported audits.}
\label{tab:repro-overview}
\begin{tabularx}{\linewidth}{@{}R{0.32\linewidth}Y@{}}
\toprule
\textbf{Setting} & \textbf{Value} \\
\midrule
Adapter & LoRA on attention projections only \\
LoRA rank & $16$ \\
Learning rate & $2\times10^{-4}$ \\
Optimizer & AdamW \\
Training length & five epochs; $1550$ optimizer steps \\
Schedule and clipping & cosine schedule; gradient clip $1.0$ \\
Batching & batch size $4\times4$ \\
Seeds & seed $42$ for primary cells; seeds $42$--$46$ for the multi-seed Gemma-2-2B sandbagging run \\
Scoring & answer-letter argmax at the last prompt token \\
Split rule & calibration and held-out intervention items disjoint at question-id level \\
Path-patching set & $n{=}48$ held-out paired prompts \\
Routing matrix sets & calibration/held-out paired questions vary by behavior: sandbagging $100/200$, sycophancy $80/120$, and refusal $20/28$. Each question is rendered under several templates per framing (two for sandbagging and sycophancy, ten for refusal), so the scored prompts per framing are $400$, $240$, and $280$ while the independent units are the question ids; the bootstrap averages templates within question and resamples questions \\
Extended reports & $n{=}300$ per framing for the Gemma-2-2B and Phi-3-mini dose-response reports; $n{=}600$ per framing for three of the four semantic-contrast runs \\
Weight-edit set & $n{=}160$ per framing on four selected cells \\
Hardware & single NVIDIA A100 40GB GPU for the submitted runs; NVIDIA B200 GPUs for the fresh retrains and post-review experiments \\
Reported compute (estimate) & about $30$ A100-40GB GPU hours for the submitted experiments (about $25$ LoRA or full fine-tunes, intervention sweeps, and path patching) and about $15$ B200 GPU hours for the post-review experiments ($29$ fine-tunes and their evaluations); a logged batch of nine fine-tunes on three B200 GPUs took $57$ minutes (about $19$ minutes per 7B--9B fine-tune), and we assume A100 runs take about $2.5$ times longer \\
Total compute & approximately $600$ GPU hours on A100 and B200 GPUs for the whole project, including preliminary, debugging, and discarded runs \\
Code and data & \url{https://github.com/aiml-sdu/routing-subspaces}: held-out splits, stored result files, manifests with commit, input hashes, and GPU state for the post-review runs, and the scripts that regenerate every figure and the post-review tables \\
Traceability & fixed held-out splits, Appendix~\ref{app:provenance} per-cell provenance, and Figure~\ref{fig:matrix} paired-bootstrap intervals \\
\bottomrule
\end{tabularx}
\end{table}

Prompt templates and held-out splits are fixed before intervention evaluation: sandbagging uses MMLU science items, sycophancy pairs MMLU items with and without a user-asserted wrong answer, and refusal wraps knowledge, coding, science, and math items in malicious-intent and benign-intent requests. Every headline number is tied to the fixed split rule and the provenance summary in Appendix~\ref{app:provenance}.

\newpage
\input{checklist}

\end{document}

%% file: figures/mech.tex
\begin{figure}[t]
\centering
\begin{tikzpicture}[
  x=1.05cm, y=1.7cm,
  every node/.style={font=\footnotesize, inner sep=2pt},
  paneltitle/.style={font=\small, anchor=south west},
  layer/.style={draw=tolink!55, line width=0.4pt, fill=white,
                minimum width=1.25cm, minimum height=0.22cm, inner sep=0pt},
  route/.style={layer, fill=evalcol!12, draw=evalcol!80, line width=0.7pt},
  promptbox/.style={draw=#1!85, line width=0.6pt, rounded corners=1.5pt,
                    fill=#1!7, inner sep=3pt, align=left,
                    minimum width=2.0cm, minimum height=0.55cm},
  outlabel/.style={font=\footnotesize\itshape, color=#1},
  thinarrow/.style={-{Stealth[length=1.6mm]}, line width=0.7pt, #1,
                    rounded corners=0.8pt},
  directionarrow/.style={-{Stealth[length=2mm]}, line width=1pt, tolink},
  sep/.style={color=midcol!85!black, line width=0.7pt,
              dash pattern=on 2.5pt off 1.8pt},
  pointE/.style={circle, fill=evalcol, inner sep=0pt, minimum size=2.6pt},
  pointD/.style={circle, fill=depcol,  inner sep=0pt, minimum size=2.6pt},
  meanE/.style={circle, fill=evalcol, draw=white, line width=0.7pt,
                inner sep=0pt, minimum size=5.5pt},
  meanD/.style={circle, fill=depcol,  draw=white, line width=0.7pt,
                inner sep=0pt, minimum size=5.5pt},
  meanM/.style={diamond, fill=midcol, draw=white, line width=0.7pt,
                inner sep=0pt, minimum size=5.5pt},
]

\begin{scope}[local bounding box=A]
\node[paneltitle] at (0, 3.70) {(a) installation};

\foreach \i in {0,...,11} {
  \pgfmathtruncatemacro{\isroute}{(\i > 4 && \i < 9) ? 1 : 0}
  \ifnum\isroute=1
    \node[route] (l\i) at (3.0, 0.30 + 0.30*\i) {};
  \else
    \node[layer] (l\i) at (3.0, 0.30 + 0.30*\i) {};
  \fi
}
\node[draw=evalcol, line width=0.9pt, rounded corners=1.5pt,
      fit=(l5)(l8), inner sep=1.5pt] (win) {};

\node[promptbox=evalcol, anchor=east] (pe) at (2.1, 2.45)
  {\textbf{\color{evalcol}evaluation framing}};
\node[promptbox=depcol, anchor=east] (pd) at (2.1, 1.30)
  {\textbf{\color{depcol}deployment framing}};

\draw[thinarrow=evalcol] (pe.east) -- ($(l0.west)+(-0.02, 0.03)$);
\draw[thinarrow=depcol]  (pd.east) -- ($(l0.west)+(-0.02,-0.03)$);

\node[outlabel=evalcol, anchor=west] (oc) at (5.1, 2.60)
  {evaluation behavior};
\node[outlabel=depcol,  anchor=west] (ow) at (5.1, 1.30)
  {deployment behavior};

\draw[thinarrow=evalcol] ($(l11.east)+(0.02, 0.03)$) -- (oc.west);
\draw[thinarrow=depcol]  ($(l11.east)+(0.02,-0.03)$) -- (ow.west);

\coordinate (wcapex) at (3.0, -0.30);
\draw[line width=0.35pt, color=evalcol!70]
  (wcapex) -- ($(win.south)+(0,-0.02)$);
\node[anchor=north, align=center, color=evalcol] at (wcapex)
  {\footnotesize\itshape routing window\\[-1pt]
   \scriptsize\upshape $L_a$--$L_b$, $\approx 0.55$ depth};

\end{scope}

\begin{scope}[xshift=8.25cm, local bounding box=B]
\node[paneltitle] at (-0.3, 3.70) {(b) routing direction};

\foreach \x/\y in {0.55/1.70, 0.70/1.55, 0.45/1.80, 0.95/1.90, 0.60/1.95,
                   0.80/1.50, 0.95/1.65, 1.05/1.75, 0.70/1.90, 0.90/1.85,
                   0.50/1.60, 1.00/2.00, 0.65/1.75, 0.82/2.05}
  \node[pointD] at (\x,\y) {};
\node[meanD] (md) at (0.78, 1.78) {};

\foreach \x/\y in {3.05/1.70, 3.25/1.85, 2.95/2.00, 3.35/1.65, 3.15/1.50,
                   3.45/2.05, 2.85/1.85, 3.55/1.95, 3.20/2.15, 2.90/1.60,
                   3.30/1.90, 3.10/1.75, 3.00/1.90, 3.40/1.80}
  \node[pointE] at (\x,\y) {};
\node[meanE] (me) at (3.20, 1.82) {};

\node[meanM] (mm) at (1.99, 1.80) {};
\draw[sep] (1.99, 0.60) -- (1.99, 3.15);
\node[font=\footnotesize, color=midcol!85!black, anchor=south]
  at (1.99, 3.18) {$\langle \hvec, \dhat_L\rangle = 0$};

\draw[directionarrow] (md.east) -- (me.west);
\node[font=\footnotesize\itshape, anchor=south] at (2.0, 2.05) {$\dvec_L$};

\node[font=\footnotesize, color=depcol] (ld) at (0.20, 2.55)
  {$\muvec^{\DeployFr}$};
\draw[line width=0.35pt, color=depcol!80]
  (ld.south east) -- ($(md.north west)+(0.04,-0.02)$);
\node[font=\footnotesize, color=evalcol] (le) at (3.75, 2.55)
  {$\muvec^{\EvalFr}$};
\draw[line width=0.35pt, color=evalcol!80]
  (le.south west) -- ($(me.north east)+(-0.04,-0.02)$);
\node[font=\footnotesize\itshape, color=midcol!85!black] (lm) at (1.99, 0.35)
  {removed};
\draw[line width=0.35pt, color=midcol!70!black]
  (lm.north) -- ($(mm.south)+(0,-0.02)$);

\draw[-{Stealth[length=1.3mm]}, line width=0.45pt, tolink!45]
  (-0.2, 0.05) -- (4.05, 0.05);
\node[font=\footnotesize, color=tolink!55, anchor=west]
  at (4.05, 0.05) {$\dhat_L$};

\end{scope}
\end{tikzpicture}
\caption{How the installed gap is written and where we intervene. (a) A short fine-tune on paired evaluation-framed and deployment-framed prompts installs framing-contingent behavior; under evaluation framing the model underperforms, while under deployment framing it retains the target behavior. Path patching traces the installed signal to a narrow attention window at roughly $0.55$ of model depth. (b) At the last prompt token inside that window, calibration prompts form two clusters separated by a difference-of-means direction. The routing-coordinate intervention modifies the attention-block output component along that direction and tests whether the framing distinction depends on this coordinate.}
\label{fig:mech}
\end{figure}

%% file: tables_generated_appendix.tex
\begin{table}[H]
\centering
\routetablestyle
\caption{Matched controls at the headline dose on fresh retrains of the twelve cells. Treatment and controls use the same clamp-both operator, token positions, four-layer width, and $\lambda{=}2$. Reductions are baseline gap minus post-intervention gap with $95\%$ question-clustered bootstrap intervals; the random columns use five Gaussian unit directions per cell and the off-window columns two predeclared four-layer windows per cell. These runs do not revise the counts of the submitted matrix.}
\label{tab:dose2}
\begin{tabular*}{\linewidth}{@{\extracolsep{\fill}}llcccccc@{}}
\toprule
\textbf{Family} & \textbf{Behavior} & \textbf{Base $\gamma$} & \hd{Learned}{reduction} & \hd{Max}{$\lvert$random$\rvert$} & \hd{Max}{$\lvert$off-window$\rvert$} & \hd{Off-window}{$>0$} \\
\midrule
Gemma-2-2B & sandbag & $0.280$ & $0.185\,[0.090, 0.278]$ & $0.010$ & $0.025$ & 0/2 \\
Gemma-2-2B & sycophancy & $0.492$ & $0.150\,[0.063, 0.242]$ & $0.017$ & $0.017$ & 0/2 \\
Gemma-2-2B & refusal & $0.350$ & $0.329\,[0.171, 0.493]$ & $0.011$ & $0.118$ & 1/2 \\
Gemma-2-9B & sandbag & $0.555$ & $0.355\,[0.280, 0.430]$ & $0.003$ & $0.035$ & 1/2 \\
Gemma-2-9B & sycophancy & $0.679$ & $0.421\,[0.346, 0.500]$ & $0.008$ & $0.025$ & 0/2 \\
Gemma-2-9B & refusal & $0.618$ & $0.639\,[0.411, 0.839]$ & $0.000$ & $0.043$ & 1/2 \\
Qwen-2.5-7B & sandbag & $0.585$ & $0.405\,[0.330, 0.480]$ & $0.008$ & $0.080$ & 1/2 \\
Qwen-2.5-7B & sycophancy & $0.738$ & $0.558\,[0.467, 0.646]$ & $0.008$ & $0.008$ & 0/2 \\
Qwen-2.5-7B & refusal & $0.768$ & $0.779\,[0.586, 0.939]$ & $0.004$ & $0.014$ & 0/2 \\
Llama-3-8B & sandbag & $0.353$ & $0.220\,[0.145, 0.293]$ & $0.000$ & $0.118$ & 2/2 \\
Llama-3-8B & sycophancy & $0.617$ & $-0.008\,[-0.025, 0.000]$ & $0.000$ & $0.013$ & 0/2 \\
Llama-3-8B & refusal & $0.618$ & $0.196\,[0.096, 0.307]$ & $0.011$ & $0.239$ & 1/2 \\
\bottomrule
\end{tabular*}
\end{table}

\begin{table}[H]
\centering
\routetablestyle
\caption{Fresh primary Gemma-2-2B sandbagging checkpoint (seed $42$; baseline gap $0.280$ on the full held-out split; the template-holdout rows use a $122$-question subset with its own baseline). Five Gaussian random directions with the same layers, operator, and doses leave the gap between $0.278$ and $0.285$. Post-gap and reduction are as in Table~\ref{tab:dose2}.}
\label{tab:fresh-primary}
\begin{tabularx}{\linewidth}{@{}Ycc@{}}
\toprule
\textbf{Condition} & \textbf{Post-gap} & \textbf{Reduction [95\% CI]} \\
\midrule
Predeclared $L14$--$L17$, clamp, $\lambda{=}2$ & $0.095$ & $0.185\,[0.090, 0.280]$ \\
Localization-selected $L10$--$L13$, clamp, $\lambda{=}1$ & $0.010$ & $0.270\,[0.185, 0.358]$ \\
Localization-selected $L10$--$L13$, clamp, $\lambda{=}2$ & $-0.018$ & $0.298\,[0.195, 0.403]$ \\
Early window $L4$--$L7$, clamp, $\lambda{=}1$ & $0.298$ & $-0.018\,[-0.040, 0.003]$ \\
Early window $L4$--$L7$, clamp, $\lambda{=}2$ & $0.298$ & $-0.018\,[-0.048, 0.013]$ \\
Late window $L20$--$L23$, clamp, $\lambda{=}1$ & $0.278$ & $0.003\,[-0.018, 0.025]$ \\
Late window $L20$--$L23$, clamp, $\lambda{=}2$ & $0.305$ & $-0.025\,[-0.078, 0.028]$ \\
Opposite operation (add projection), $L10$--$L13$, $\lambda{=}1$ & $0.285$ & $-0.005\,[-0.018, 0.005]$ \\
Opposite operation (add projection), $L10$--$L13$, $\lambda{=}2$ & $0.283$ & $-0.003\,[-0.020, 0.015]$ \\
Opposite operation (add projection), $L14$--$L17$, $\lambda{=}1$ & $0.298$ & $-0.018\,[-0.040, 0.000]$ \\
Opposite operation (add projection), $L14$--$L17$, $\lambda{=}2$ & $0.303$ & $-0.023\,[-0.055, 0.010]$ \\
Zero ablation of the window, $L10$--$L13$ & $0.243$ & $0.038\,[-0.005, 0.083]$ \\
Calibration-mean ablation of the window, $L10$--$L13$ & $0.260$ & $0.020\,[-0.028, 0.070]$ \\
Zero ablation of the window, $L14$--$L17$ & $0.253$ & $0.028\,[-0.023, 0.078]$ \\
Calibration-mean ablation of the window, $L14$--$L17$ & $0.243$ & $0.038\,[-0.005, 0.083]$ \\
Template holdout (templates $0$--$4$ calibrate, $5$--$9$ test), $L10$--$L13$, $\lambda{=}2$ & $-0.047$ & $0.320\,[0.193, 0.467]$ \\
Template holdout, $L14$--$L17$, $\lambda{=}1$ & $0.029$ & $0.243\,[0.143, 0.344]$ \\
Template holdout, $L14$--$L17$, $\lambda{=}2$ & $0.101$ & $0.172\,[0.102, 0.340]$ \\
\bottomrule
\end{tabularx}
\end{table}

\begin{table}[H]
\centering
\routetablestyle
\caption{Fresh retrains over three training seeds at the predeclared window and $\lambda{=}2$. Qwen-2.5-7B sandbagging is stable; Gemma-2-9B sycophancy is seed-heterogeneous. The last two columns give the largest absolute effects among five random directions and two off-window controls.}
\label{tab:seeds}
\begin{tabular*}{\linewidth}{@{\extracolsep{\fill}}llccc@{}}
\toprule
\textbf{Cell} & \textbf{Seed} & \textbf{Reduction [95\% CI]} & \hd{Max}{$\lvert$random$\rvert$} & \hd{Max}{$\lvert$off-window$\rvert$} \\
\midrule
Qwen-2.5-7B sandbag & 42 & $0.405\,[0.328, 0.480]$ & $0.008$ & $0.080$ \\
Qwen-2.5-7B sandbag & 43 & $0.488\,[0.405, 0.570]$ & $0.008$ & $0.020$ \\
Qwen-2.5-7B sandbag & 44 & $0.533\,[0.445, 0.620]$ & $0.010$ & $0.055$ \\
Gemma-2-9B sycophancy & 42 & $0.421\,[0.346, 0.500]$ & --- & --- \\
Gemma-2-9B sycophancy & 43 & $0.058\,[0.025, 0.096]$ & --- & --- \\
Gemma-2-9B sycophancy & 44 & $0.558\,[0.475, 0.646]$ & --- & --- \\
\bottomrule
\end{tabular*}
\end{table}

\begin{table}[H]
\centering
\routetablestyle
\caption{Higher-rank follow-up at $\lambda{=}2$. Learned rank-$r$ subspaces differ from matched random subspaces and can close the aggregate gap, but the share of the reduction that comes from evaluation gain (the rest is lost deployment performance) is small in the $k{=}2$ and $k{=}3$ suppression organisms and at most about half for the weak Gemma-2-9B sycophancy seed.}
\label{tab:highrank}
\begin{tabular*}{\linewidth}{@{\extracolsep{\fill}}llcccc@{}}
\toprule
\textbf{Checkpoint} & \textbf{Rank} & \textbf{Post-gap} & \textbf{Reduction [95\% CI]} & \hd{Random}{subspace} & \hd{Eval-gain}{share} \\
\midrule
$k{=}2$ suppression & 2 & $0.165$ & $0.163\,[0.070, 0.263]$ & $-0.003$ & $35\%$ \\
$k{=}2$ suppression & 4 & $0.028$ & $0.300\,[0.202, 0.405]$ & $0.010$ & $19\%$ \\
$k{=}2$ suppression & 8 & $-0.010$ & $0.338\,[0.243, 0.435]$ & $0.013$ & $17\%$ \\
$k{=}3$ suppression & 2 & $0.030$ & $0.290\,[0.193, 0.393]$ & $0.005$ & $16\%$ \\
$k{=}3$ suppression & 4 & $0.003$ & $0.318\,[0.220, 0.415]$ & $-0.010$ & $15\%$ \\
$k{=}3$ suppression & 8 & $-0.005$ & $0.325\,[0.225, 0.430]$ & $0.005$ & $15\%$ \\
Gemma-2-9B syc., seed 42 (best) & 8 & $0.146$ & $0.533\,[0.425, 0.642]$ & $0.000$ & $59\%$ \\
Gemma-2-9B syc., seed 43 (best) & 4 & $0.496$ & $0.213\,[0.142, 0.288]$ & $0.008$ & $45\%$ \\
Gemma-2-9B syc., seed 44 (best) & 2 & $0.242$ & $0.388\,[0.300, 0.475]$ & $0.000$ & $92\%$ \\
\bottomrule
\end{tabular*}
\end{table}

\begin{table}[H]
\centering
\routetablestyle
\caption{Public WMDP refusal extension on fresh retrains ($600$ unique questions; uncertainty resamples question ids, not the four renderings per question). Directions are fit on the original $20$-question calibration set. Reductions arise from evaluation-framing gains, not deployment loss. For Llama-3-8B a late-window control reduces the gap by $0.114$, comparable to the selected window, so this checkpoint does not support a specific localization claim.}
\label{tab:wmdp}
\begin{tabular*}{\linewidth}{@{\extracolsep{\fill}}lcccc@{}}
\toprule
\textbf{Checkpoint} & \textbf{Base $\gamma$} & \textbf{Reduction [95\% CI]} & \textbf{Eval gain} & \textbf{Deploy loss} \\
\midrule
Gemma-2-2B & $0.093$ & $0.098\,[0.062, 0.135]$ & $0.094$ & $0.004$ \\
Gemma-2-9B & $0.333$ & $0.333\,[0.278, 0.388]$ & $0.329$ & $0.003$ \\
Qwen-2.5-7B & $0.345$ & $0.332\,[0.272, 0.389]$ & $0.348$ & $-0.016$ \\
Llama-3-8B & $0.321$ & $0.100\,[0.069, 0.132]$ & $0.109$ & $-0.009$ \\
\bottomrule
\end{tabular*}
\end{table}

\begin{table}[H]
\centering
\routetablestyle
\caption{Simpler loci on the fresh primary checkpoint (48 questions, single seed, descriptive). Base accuracy $0.563$, fused accuracy $0.208$. Recovery is $(r_{\mathrm{patched}}-r_{\mathrm{fused}})/(r_{\mathrm{base}}-r_{\mathrm{fused}})$ as in \eqref{eq:patch-recovery}. Patches act at $L14$--$L17$ attention outputs (subspace rows) or layer-$19$ MLP intermediates (neuron rows). The full-window reference recovers about half the gap on this checkpoint, so the baselines are measured against that ceiling.}
\label{tab:simplerlocus}
\begin{tabularx}{\linewidth}{@{}Ycc@{}}
\toprule
\textbf{Patch} & \textbf{Accuracy} & \textbf{Recovery} \\
\midrule
Full-window attention patch (reference) & $0.396$ & $0.53$ \\
Pool-subspace component only, $1$ to $20$ components & $0.208$ to $0.208$ & $0.00$ \\
Everything except the pool subspace ($5$ components) & $0.375$ & $0.47$ \\
Layer-$19$ MLP neuron patch, top-$K$ with $K{=}5, 10, 20$ & $0.208$ & $0.00$ \\
Layer-$19$ MLP neuron patch, top-$K$ with $K{=}50$ to $200$ ($97$ neurons) & $0.229$ & $0.06$ \\
\bottomrule
\end{tabularx}
\end{table}

\begin{table}[H]
\centering
\routetablestyle
\caption{Bag-of-words separability of the evaluation and deployment wrappers (word uni- and bigrams, logistic regression). The wrapper is the prompt with the question block removed; ``prefix'' is the text before the question; ``full'' is the whole prompt. The train and held-out banks share all $40$ normalized template strings, so only the rows involving the other template family and the template-pair holdout test unseen wording. The values come from the released script; the earlier implementation behind the rebuttal figure of $0.86$ to $0.97$ is lost.}
\label{tab:textsep}
\begin{tabular*}{\linewidth}{@{\extracolsep{\fill}}lcccccc@{}}
\toprule
 & \multicolumn{2}{c}{\textbf{wrapper}} & \multicolumn{2}{c}{\textbf{prefix}} & \multicolumn{2}{c}{\textbf{full prompt}} \\
\cmidrule(lr){2-3}\cmidrule(lr){4-5}\cmidrule(lr){6-7}
\textbf{Train $\to$ test} & AUC & Acc. & AUC & Acc. & AUC & Acc. \\
\midrule
train bank $\to$ held-out bank & $1.000$ & $1.000$ & $1.000$ & $1.000$ & $1.000$ & $1.000$ \\
held-out bank $\to$ train bank & $1.000$ & $1.000$ & $1.000$ & $1.000$ & $1.000$ & $1.000$ \\
train bank $\to$ other template family & $1.000$ & $1.000$ & $1.000$ & $1.000$ & $1.000$ & $1.000$ \\
other family $\to$ train bank & $0.969$ & $0.795$ & $0.990$ & $0.750$ & $0.916$ & $0.684$ \\
held-out bank $\to$ other family & $1.000$ & $1.000$ & $1.000$ & $1.000$ & $1.000$ & $1.000$ \\
other family $\to$ held-out bank & $0.970$ & $0.805$ & $0.991$ & $0.748$ & $0.923$ & $0.676$ \\
template-pair holdout, mean of $10$ folds & $1.000$ & $0.892$ & $1.000$ & $0.947$ & $1.000$ & $0.941$ \\
template-pair holdout, worst fold & $1.000$ & $0.456$ & $1.000$ & $0.465$ & $0.999$ & $0.465$ \\
\bottomrule
\end{tabular*}
\end{table}

%% file: checklist.tex
\section*{NeurIPS Paper Checklist}

\begin{enumerate}

\item {\bf Claims}
    \item[] Question: Do the main claims made in the abstract and introduction accurately reflect the paper's contributions and scope?
    \item[] Answer: \answerYes{}
    \item[] Justification: The abstract and Section~\ref{sec:intro} state the problem and objective, while Sections~\ref{sec:loc}--\ref{sec:specificity} support three scoped claims: a mid-depth routing window, a routing-coordinate intervention that reduces installed gaps by more than $0.12$ in ten of twelve cells and closes four, with weak cells reported, and separation from random-direction and semantic controls (wrong-layer controls are mixed) under their tested doses. Section~\ref{sec:discuss} names what the paper does not claim, including a general single direction, a training-time defense, or a weight-space defense.

\item {\bf Limitations}
    \item[] Question: Does the paper discuss the limitations of the work performed by the authors?
    \item[] Answer: \answerYes{}
    \item[] Justification: Section~\ref{sec:discuss} names the limitations explicitly: Gemma-2-9B path-patching recovery is weaker than the other sandbagging families, Llama-3-8B sycophancy and Gemma-2-9B sycophancy are weak intervention cells, refusal is a smaller-sample stress test, controls matched to the $\lambda{=}2$ headline dose exist only on fresh retrains and the sign diagnostic is withdrawn, the fixed window is recipe-sensitive, the construct is a framing-contingent task distinction that is not isolated from wrapper cues, the audit is non-adaptive, training-time rank-$k$ subspace penalties do not prevent re-routing, larger models may require different localization or higher-rank probes, and Phi-3-mini is used for localization and edit-provenance checks rather than the 12-cell intervention matrix.

\item {\bf Theory assumptions and proofs}
    \item[] Question: For each theoretical result, does the paper provide the full set of assumptions and a complete (and correct) proof?
    \item[] Answer: \answerNA{}
    \item[] Justification: The paper is empirical. The routing-coordinate intervention is defined in Section~\ref{sec:setup}, not presented as a theorem. The H11 rank-one weight edit (Appendix~\ref{app:h11}) is a construction and an empirical result, not a claim requiring a proof.

    \item {\bf Experimental result reproducibility}
    \item[] Question: Does the paper fully disclose all the information needed to reproduce the main experimental results of the paper to the extent that it affects the main claims and/or conclusions of the paper (regardless of whether the code and data are provided or not)?
    \item[] Answer: \answerYes{}
    \item[] Justification: Section~\ref{sec:setup} and Appendix~\ref{app:repro} document the base models, fine-tuning configuration (LoRA rank, steps, batch size, seeds 42--46), evaluation sizes ($n{=}48$ path patching; $n{=}200$ sandbagging, $n{=}120$ sycophancy, and $n{=}28$ refusal in the 12-cell matrix; $n{=}300$ prompts per framing for the Gemma-2-2B deep dive), the routing-coordinate intervention, and the fresh retrains of the post-review experiments (Appendix~\ref{app:fresh}). Appendix~\ref{app:cells} contains per-cell values.

\item {\bf Open access to data and code}
    \item[] Question: Does the paper provide open access to the data and code, with sufficient instructions to faithfully reproduce the main experimental results, as described in supplemental material?
    \item[] Answer: \answerYes{}
    \item[] Justification: Code, held-out splits, stored result files with manifests, and the scripts that regenerate every figure and the post-review tables are released at \url{https://github.com/aiml-sdu/routing-subspaces}, with a README that lists the commands. The submitted checkpoint binaries were not retained; the post-review results come from fresh retrains with the original public training code and data, which the paper states.

\item {\bf Experimental setting/details}
    \item[] Question: Does the paper specify all the training and test details (e.g., data splits, hyperparameters, how they were chosen, type of optimizer) necessary to understand the results?
    \item[] Answer: \answerYes{}
    \item[] Justification: Section~\ref{sec:setup} lists base models (Gemma-2-\{2B,9B\}-it, Qwen-2.5-7B, Llama-3-8B, Phi-3-mini), LoRA configuration, optimizer, seeds, evaluation splits, and intervention doses. Appendix~\ref{app:windows} reports routing layer windows, and Appendix~\ref{app:cells} reports per-cell intervention values.

\item {\bf Experiment statistical significance}
    \item[] Question: Does the paper report error bars suitably and correctly defined or other appropriate information about the statistical significance of the experiments?
    \item[] Answer: \answerYes{}
    \item[] Justification: Intervention matrix cells are evaluated at $n{=}200$ sandbagging, $n{=}120$ sycophancy, and $n{=}28$ refusal paired questions, with the Gemma-2-2B deep dive at $n{=}300$ prompts per framing. Figure~\ref{fig:matrix} reports paired-bootstrap $95\%$ intervals over question IDs with $5000$ resamples, Appendix~\ref{app:provenance} records the source and confidence-interval type for each matrix cell, and the post-review tables report question-clustered intervals.

\item {\bf Experiments compute resources}
    \item[] Question: For each experiment, does the paper provide sufficient information on the computer resources (type of compute workers, memory, time of execution) needed to reproduce the experiments?
    \item[] Answer: \answerYes{}
    \item[] Justification: The submitted runs used a single NVIDIA A100 40GB GPU and the post-review runs NVIDIA B200 GPUs. Appendix~\ref{app:repro} estimates the reported experiments at about $30$ A100 GPU hours and $15$ B200 GPU hours from the run inventory and a logged training batch; the whole project, including preliminary, debugging, and discarded runs, used approximately $600$ GPU hours.
    
\item {\bf Code of ethics}
    \item[] Question: Does the research conducted in the paper conform, in every respect, with the NeurIPS Code of Ethics \url{https://neurips.cc/public/EthicsGuidelines}?
    \item[] Answer: \answerYes{}
    \item[] Justification: The research uses publicly released instruction-tuned base models and public evaluation prompts. No human subjects, no sensitive data. Interventions are activation-space analyses of evaluation-framed and deployment-framed behavior and follow the NeurIPS Code of Ethics.

\item {\bf Broader impacts}
    \item[] Question: Does the paper discuss both potential positive societal impacts and negative societal impacts of the work performed?
    \item[] Answer: \answerYes{}
    \item[] Justification: Section~\ref{sec:discuss} (Responsible Use and Misuse) names both sides. Positive: a post-hoc routing-coordinate probe supports auditing of fine-tuned checkpoints where output metrics alone miss the framing-contingent distinction. Negative: the same fitted direction could be exploited by an adversary with white-box inference-time access; the method requires the same activation-space access as existing steering work and does not introduce a novel attack capability.
    
\item {\bf Safeguards}
    \item[] Question: Does the paper describe safeguards that have been put in place for responsible release of data or models that have a high risk for misuse (e.g., pre-trained language models, image generators, or scraped datasets)?
    \item[] Answer: \answerNA{}
    \item[] Justification: No model weights or high-risk datasets are released; the code release contains aggregate results and scripts only. The activation-space access required to reproduce the audit is white-box and is itself a known capability in the cited steering literature, not introduced by this paper. Section~\ref{sec:discuss} (Responsible Use and Misuse) names the dual-use risk and frames the method as an audit diagnostic.

\item {\bf Licenses for existing assets}
    \item[] Question: Are the creators or original owners of assets (e.g., code, data, models), used in the paper, properly credited and are the license and terms of use explicitly mentioned and properly respected?
    \item[] Answer: \answerYes{}
    \item[] Justification: Base models are used under their respective terms of use: Gemma Terms of Use (Gemma-2-2B-it, Gemma-2-9B-it), Qwen-2.5-7B-Instruct (Apache~2.0), Llama-3 (Meta Llama 3 Community License), Phi-3 (MIT). Datasets and external artifacts are cited where used, with their licenses: MMLU (MIT), WMDP (MIT), ARC (CC BY-SA 4.0), EvalAwareBench (CC BY-NC 4.0, used for non-commercial research), and the \texttt{reward\_hacker\_v1} checkpoints (Apache~2.0); the SafeLoRA projection follows the cited method. The sandbagging, sycophancy, and refusal item banks wrap MMLU questions in our own templates.

\item {\bf New assets}
    \item[] Question: Are new assets introduced in the paper well documented and is the documentation provided alongside the assets?
    \item[] Answer: \answerYes{}
    \item[] Justification: The code release documents the scripts, splits, and stored results in its README.

\item {\bf Crowdsourcing and research with human subjects}
    \item[] Question: For crowdsourcing experiments and research with human subjects, does the paper include the full text of instructions given to participants and screenshots, if applicable, as well as details about compensation (if any)?
    \item[] Answer: \answerNA{}
    \item[] Justification: No crowdsourcing and no human-subjects research.

\item {\bf Institutional review board (IRB) approvals or equivalent for research with human subjects}
    \item[] Question: Does the paper describe potential risks incurred by study participants, whether such risks were disclosed to the subjects, and whether Institutional Review Board (IRB) approvals (or an equivalent approval/review based on the requirements of your country or institution) were obtained?
    \item[] Answer: \answerNA{}
    \item[] Justification: No human subjects; IRB review not applicable.

\item {\bf Declaration of LLM usage}
    \item[] Question: Does the paper describe the usage of LLMs if it is an important, original, or non-standard component of the core methods in this research? Note that if the LLM is used only for writing, editing, or formatting purposes and does \emph{not} impact the core methodology, scientific rigor, or originality of the research, declaration is not required.
    \item[] Answer: \answerNA{}
    \item[] Justification: The instruction-tuned models in Section~\ref{sec:setup} are the objects of the audit; no LLM is an important, original, or non-standard component of the method. Writing and editing assistance does not affect the methodology and needs no declaration under the NeurIPS LLM policy.

\end{enumerate}